\documentclass[10pt,journal,compsoc]{IEEEtran}

\ifCLASSOPTIONcompsoc
  \usepackage[nocompress]{cite}
\else
  \usepackage{cite}
\fi

\ifCLASSINFOpdf
   \usepackage[pdftex]{graphicx}
   \graphicspath{{./figures/}}
    \DeclareGraphicsExtensions{.jpg,.jpeg,.pdf,.png} 
\else
  
\fi

\usepackage{dsfont}
\usepackage{amsmath}
\usepackage{amssymb}
\usepackage{bbm}

 \usepackage[caption=false,font=footnotesize,labelfont=sf,textfont=sf]{subfig}

\usepackage[usenames, dvipsnames]{color}
\usepackage[dvipsnames]{xcolor}
\usepackage[hidelinks]{hyperref}
\usepackage{enumitem}

\usepackage{soul}

\usepackage{setspace}
\usepackage{tabu} 

\usepackage{times}
\usepackage{epsfig}
\usepackage{graphicx}
\usepackage{amsmath}
\usepackage{amssymb}
\usepackage{xcolor}

\usepackage{multirow}
\usepackage{amsthm,amsmath,amssymb}
\usepackage{mathrsfs}
\usepackage{graphicx}
\usepackage{amsmath}
\usepackage{amssymb}
\usepackage{booktabs}
\usepackage{times}
\usepackage{microtype}
\usepackage{epsfig}
\usepackage{placeins}
\newcommand{\et}[2]{${#1}^{\pm{#2}}$}

\newcommand{\etr}[2]{$\boldsymbol{{#1}}^{\pm{#2}}$}
\newcommand{\etbb}[2]{$\underline{{#1}}^{\pm{#2}}$}

\newcommand{\revise}[1]{\textcolor{black}{#1}}
\newif\ifcomments
\commentstrue 

\newcommand{\commentZY}[1]{\textcolor{black}{#1}}
\newcommand{\ywu}[1]{\textcolor{black}{#1}}

\usepackage[switch]{lineno}

\begin{document}
%
\title{GUESS: GradUally Enriching SyntheSis \\ for Text-Driven Human Motion Generation}
%

%
%

\author{Xuehao Gao,
        Yang Yang,
        Zhenyu Xie, 
        Shaoyi Du,
        Zhongqian Sun,
        and Yang Wu
\IEEEcompsocitemizethanks{\IEEEcompsocthanksitem X. Gao and S. Du are with the National Key Laboratory of Human-Machine Hybrid Augmented Intelligence, Xi’an Jiaotong University, Xi’an 710049, China (e-mail: gaoxuehao.xjtu@gmail.com; dushaoyi@gmail.com).
\IEEEcompsocthanksitem Y. Yang is with the School of Electronic and Information Engineering, Xi’an Jiaotong University, Xi’an 710049, China (e-mail: yyang@mail.xjtu.edu.cn).
\IEEEcompsocthanksitem Z. Xie is with the School of Intelligent Systems Engineering, Sun Yat-sen University, GuangZhou 510275, China (e-mail: xiezhy6@mail2.sysu.edu.cn).
\IEEEcompsocthanksitem Z. Sun, and Y. Wu are with Tencent AI Lab. Shenzhen 518057, China (e-mail: \{sallensun, dylanywu\}@tencent.com).}
}
\IEEEtitleabstractindextext{
	\begin{abstract}

		\ywu{In this paper, we propose a novel cascaded diffusion-based generative framework for text-driven human motion synthesis, which exploits a strategy named \textbf{G}rad\textbf{U}ally \textbf{E}nriching \textbf{S}ynthe\textbf{S}is (\textbf{GUESS} as its abbreviation). The strategy sets up generation objectives by grouping body joints of detailed skeletons in close semantic proximity together and then replacing each of such joint group with a single body-part node. Such an operation recursively abstracts a human pose to coarser and coarser skeletons at multiple granularity levels. \revise{With gradually increasing the abstraction level, human motion becomes more and more concise and stable, significantly benefiting the cross-modal motion synthesis task. The whole text-driven human motion synthesis problem is then divided into multiple abstraction levels and solved with a multi-stage generation framework with a cascaded latent diffusion model: an initial generator first generates the coarsest human motion guess from a given text description; then, a series of successive generators gradually enrich the motion details based on the textual description and the previous synthesized results.} Notably, we further integrate GUESS with the proposed dynamic multi-condition fusion mechanism to dynamically balance the cooperative effects of the given textual condition and synthesized coarse motion prompt in different generation stages. Extensive experiments on large-scale datasets verify that GUESS outperforms existing state-of-the-art methods by large margins in terms of accuracy, realisticness, and diversity. Our is available at \href{https://github.com/Xuehao-Gao/GUESS}{https://github.com/Xuehao -Gao/GUESS}.}
	\end{abstract}
\vspace{-0.1in}
	}


\maketitle

\section{Introduction}
\IEEEPARstart{A}{s} a fundamental yet challenging task in computer animation, human motion synthesis brings broad applications into the real world. Given various control signals, such as natural language description \cite{DBLP:journals/tvcg/FanXG12,DBLP:conf/eccv/PetrovichBV22,DBLP:conf/eccv/GuoZWC22,DBLP:journals/corr/abs-2304-12571,xie2023b2ahdm}, voice or music audio \cite{DBLP:conf/iccv/LiYRK21,DBLP:conf/aaai/LiZZS22,DBLP:conf/nips/LeeY0WLYK19,DBLP:journals/corr/abs-2111-12159,DBLP:journals/tvcg/FanXG12}, a synthesis algorithm enables a machine to generate realistic and diverse 3D human motions from these condition inputs, benefiting VR content design, game and film creation, etc \cite{DBLP:journals/tvcg/BoukhaymaB19,yang2022motion,gao2021efficient, gao2021contrastive, DBLP:journals/tvcg/KwonCPS08,DBLP:journals/tvcg/WangCX21}. Its overall strategy is to learn a powerful cross-modal mapping function that effectively converts the latent distribution of control commands into the human motion domain. As a natural descriptor, language-based textual control signals are convenient for users to interact with motion synthesis systems, \commentZY{making text-driven motion synthesis an increasingly popular direction in the visualization and computer graphics community} \cite{DBLP:conf/3dim/AhujaM19,gao2023glimpse, DBLP:journals/corr/abs-2208-15001,DBLP:journals/corr/abs-2212-08526,gao2023decompose,gao2023learning,DBLP:journals/corr/abs-2209-14916}. 
\commentZY{However, existing methods neglect the structure of the human body and employ the generative model to convert the input control signal to the intact human motion with detailed body joints. Due to the huge discrepancy between the textual modality and motion modality, such direct conversion (i.e., transferring the text input into the intact motion sequence) \ywu{faces a great} difficulty of cross-model learning.} 

\begin{figure}[!t]
	\centering
	\includegraphics[width=0.49\textwidth]{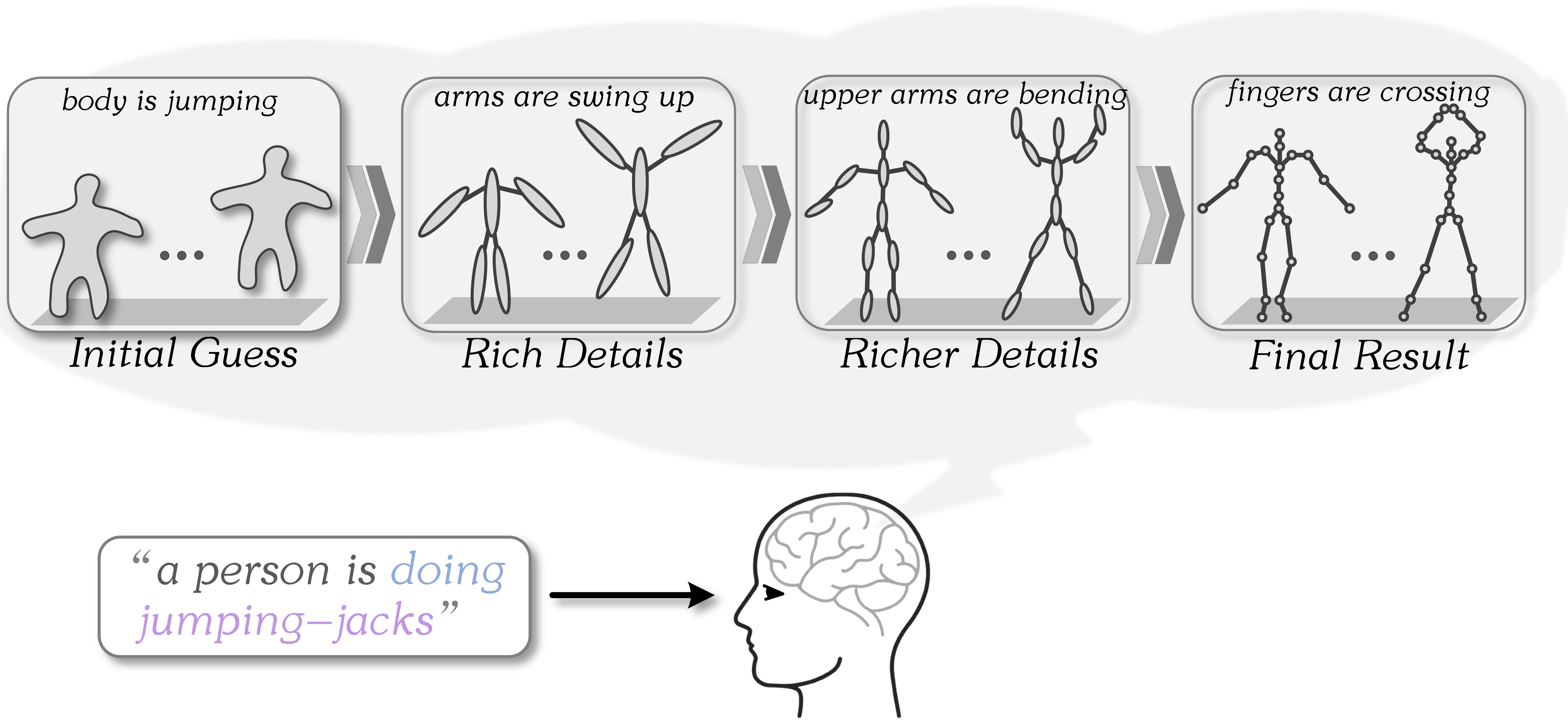}  
	\vspace{-0.25in}
	\caption{Given a textual description, the human brain imagines \ywu{the corresponding} motion visualization by inferring the body poses at the coarse body part level first and then enriching \ywu{the} finer motion details gradually.}
	\label{intro}
\end{figure}

\ywu{N}europsychology studies \ywu{show that when} given a textual description \ywu{of human motions}, the human brain tends to \ywu{imagine the corresponding visual sequence in} a coarse-to-fine \ywu{way} \cite{DBLP:journals/nature,BRUNEC20182129,memory2017}. As sketched in Figure \ref{intro}, \ywu{the brain probably} first vaguely \ywu{imagines} the coarse trajectory-related cues of the human pose sequence from its text description. Then, based on this initial guess, \ywu{it} further gradually enrich \ywu{the} motion details of body parts and body joints\ywu{, again} with the guidance of textual descriptions. 
\commentZY{Inspired by this observation, we explore a novel \textbf{G}rad\textbf{U}ally \textbf{E}nriching \textbf{S}ynthe\textbf{S}is strategy \ywu{with} a cascaded diffusion-based framework \ywu{for its realization}, named \textbf{GUESS}, to progressively utilize recursively abstracted human motion sequences as intermediate prompts for coarse-to-fine text-driven motion synthesis.} \ywu{Such intermediate prompts bring extra and varying guidance so that the proposed cascaded diffusion can generate higher-quality and more diverse results than straightforward single-stage diffusion.}



\begin{figure*}[t]
	\centering
	\includegraphics[width=0.99\textwidth]{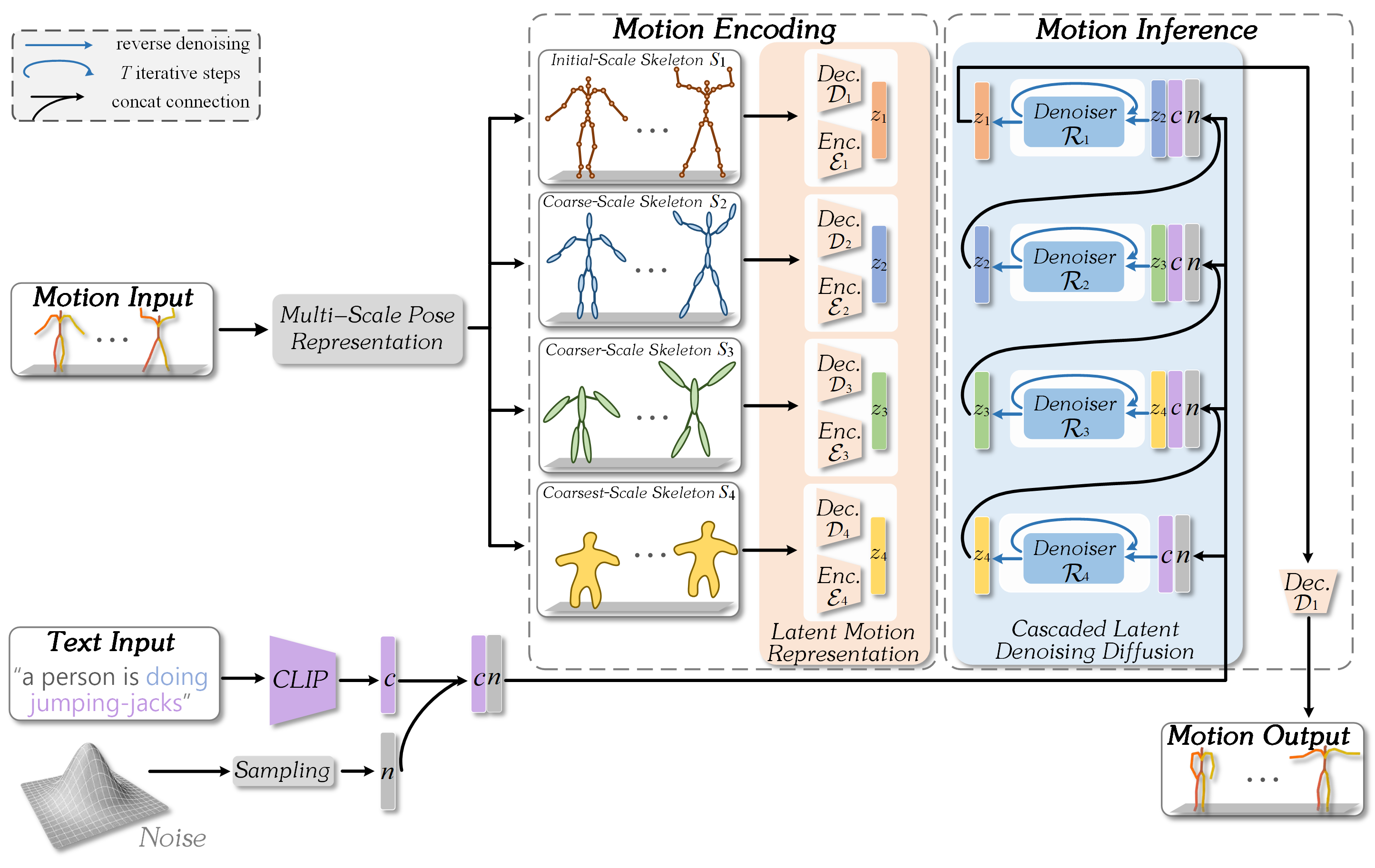}  
	\vspace{-0.15in}
	\caption{\ywu{The} Framework \ywu{of GUESS}. In the training \ywu{phase}, we first represent a human motion input with multiple pose scales ($\boldsymbol{S}_{1},\cdots,\boldsymbol{S}_{4}$) and train a motion embedding module to learn an effective latent motion representation ($z_{1},\cdots,z_{4}$) within each scale. Then, we train a cascaded latent-based diffusion model to learn a powerful probabilistic text-to-motion mapping with a joint guidance of gradually richer motion embedding ($z_{4}, \cdots, z_{2}$) and textual condition embedding ($c$). In the test \ywu{phase}, the motion inference module generates the motion embedding of the finest pose sale ($z_{1}$) from the text embedding and sends it to the corresponding decoder ($\mathcal{D}_{1}$) for the 3D motion reconstruction.}
	\label{pipeline}
\end{figure*}


\commentZY{Specially, in our novel generation strategy of gradually enriching synthesis, the detailed human skeleton is abstracted by grouping body joints in close semantic proximity together and representing each joint group with a body-part node, yielding a coarser pose. Since these body joint groups can be abstracted into body parts at different semantic granularity levels, we represent a human pose with a multi-scale skeletal graph, whose nodes are body components at various scales and edges are pairwise relations between these components. \revise{Notably, with gradually increasing abstraction levels, the body motion becomes more concise and more stable, which significantly benefits the cross-modal motion synthesis task.} In particular, suppose the pose abstraction is developed to its extremes and all body joints are grouped into a single body-level node, this coarsest structure can only present the whole-body trajectory, $i.e.,$ pose-insensitive location information.}



\revise{Benefiting from richer condition contexts and simpler synthesis targets, GUESS significantly improves the quality of human motions generated from its each text-to-motion synthesis stage. Specifically, via dividing text-to-motion synthesis into multiple abstraction levels, the coarser motion inferred from the former generator serves as an initial guess for progressively enriching the details at the next text-to-motion synthesis step based on the given textual description.} As shown in Figure \ref{pipeline}, GUESS contains two core modules: \textit{Motion Encoding} and \textit{Motion Inference}. Firstly, the \textit{Motion Encoding Module} deploys a variational autoencoder on each pose scale to learn its low-dimensional latent motion representation. Then, the \textit{Motion Inference Module} utilizes a cascaded latent diffusion model to progressively generate the target motion representation conditioning on the given CLIP-encoded text description, and inferred intermediate coarser motion guesses from the previous stage.

Furthermore, considering text description and motion guess reflect different cues inside the joint condition for generation, we thus dynamically infer adaptive condition weights of the given textual condition and the synthesized coarser motion prompt in each input sample and generation stage. Specifically, the condition weights of the given textual condition and the synthesized coarser motion prompt are sample-dependent and adaptively inferred over different denoising steps within the cascaded latent diffusion model. Coupling GUESS with this dynamic multi-condition fusion scheme, we propose a powerful text-driven human motion synthesis system that outperforms state-of-the-art methods in terms of accuracy, realisticness, and diversity.   
The main contributions of this paper are:
\begin{itemize}
	\item We explore a novel strategy of gradually enriching synthesis (GUESS) for the text-driven motion generation task. Based on this strategy, we propose a cascaded latent diffusion network to learn an effective text-to-motion mapping with cooperative guidance of textual condition embedding and gradually richer motion embedding.        
	\item We propose a dynamic multi-condition fusion mechanism that adaptively infers joint conditions of the given textual description and the synthesized coarser motion guess in each input sample and its different generation stages, significantly improving the latent-based diffusion model.  
	\item Integrating GUESS with the adaptive multi-condition fusion, we develop a powerful text-to-action generation system that outperforms state-of-the-art methods by a large margin in accuracy, realism, and diversity.
\end{itemize}

\section{Related Work}
\subsection{Text-driven Human Motion Synthesis}
Intuitively, text-driven human motion synthesis can be regarded as a text-to-motion translation task. Notably, the inherent many-to-many problem behind this task makes generating realistic and diverse human motions \ywu{very} challenging. For example, the action \ywu{described by the same word `running'} can refer to different running speeds, paths, and styles. Meanwhile, we can describe a specific human motion sample with different words. Recently, many works on this task have made great efforts and fruitful progress\ywu{es}. Specifically, JL2P \cite{DBLP:conf/3dim/AhujaM19} learns text-to-motion cross-modality mapping with a Variational AutoEncoder (VAE), suffering from the one-to-one mapping limitation. T2M \cite{DBLP:conf/cvpr/GuoZZ0JL022} engages a temporal VAE framework to extract motion snippet codes and samples latent vectors from them for human motion reconstruction. Similarly, TEMOS \cite{DBLP:conf/eccv/PetrovichBV22} proposes a VAE-based architecture to learn a joint latent space of motion and text constrained on a Gaussian distribution. However, we propose to improve the text-driven human motion synthesis task with a multi-stage generation strategy, which gradually enriches its inference result with the textual description and coarser motion prompt. As an initial attempt to explore such a coarse-to-fine generation scheme, we hope it will inspire more investigation and exploration in the community.

\subsection{Conditional Diffusion Models}
As an emerging yet promising generative framework, diffusion model significantly promotes the development of many \ywu{research} fields and brings various real-world applications, such as Imagen \cite{saharia2022photorealistic}, Dall2 \cite{ramesh2022hierarchical}, and ChatGPT \cite{chatgpt}. Inspired by the stochastic diffusion process in Thermodynamics, a sample from the target distribution is gradually noised by the diffusion process \cite{DBLP:conf/icml/Sohl-DicksteinW15,DBLP:conf/nips/0011E20,DBLP:journals/corr/abs-2209-00796}. Then, a neural diffusion model learns the reverse process from denoising the sample step by step. Encouraged by the fruitful developments of the diffusion model, \ywu{a few} recent works incorporate conditional denoising diffusion probabilistic models into the text-driven motion synthesis task, such as MotionDiffuse \cite{DBLP:journals/corr/abs-2208-15001}, MDM \cite{DBLP:journals/corr/abs-2209-14916}, MLD \cite{DBLP:journals/corr/abs-2212-04048}, and MoFusion \cite{DBLP:journals/corr/abs-2212-04495}. However, these diffusion models focus on reconstructing the distribution of human motion from a noise signal \ywu{using} a text-only description condition. In this paper, we propose a powerful cascaded latent diffusion model which facilitates the probabilistic text-to-motion mapping with cooperative guidance of textual description embedding and gradually richer motion embedding. 

\subsection{Progressive Generative Models}

Our other relevant works are progressive generative models. In the study of image generation, many methods adopt a progressive generation scheme to generate images of increasing resolution by inferring a low-resolution guess first and then successively adding higher-resolution details. Specifically, imagen video \cite{DBLP:journals/corr/abs-2210-02303} and CDM \cite{ho2022cascaded} generate high-definition images with a base low-resolution generator and a sequence of interleaved super-resolution generators. These successful attempts at image generation verify the effectiveness of the progressive generation strategy. \ywu{In contrast to} fruitful progressive image generation \ywu{attempts}, \ywu{progressive} generation scheme for 3D body pose synthesis remains \ywu{unexplored}. To the best of our knowledge, GUESS takes the first and inspiring step toward the coarse-to-fine progressive cross-modal human motion generation.  

\section{Problem Formulation}
The goal of our text-driven human motion synthesis system is to develop a powerful generator $\mathcal{F}_{\text {gen}}$ that synthesizes a realistic body pose sequence $\boldsymbol{S}$ from a natural language description or action class label condition $\boldsymbol{C}$ as $\boldsymbol{S}=\mathcal{F}_{\text {gen}}(\boldsymbol{C})$. Given a ground truth text-motion input pair, we deploy a pre-trained CLIP \cite{DBLP:conf/icml/RadfordKHRGASAM21} as a feature extractor to extract a $C_{e}$-dimensional condition embedding $c\in \mathbb{R}^{C_{e}}$ from the text input. As for the human motion input $\boldsymbol{S}$, its initial motion features \commentZY{inherit the widely-used motion representation in~\cite{DBLP:conf/cvpr/GuoZZ0JL022}, which} include 3D joint rotations, positions, velocities, and foot contact information.    

\section{Method}
\label{method}
As shown in Figure \ref{pipeline}, GUESS contains three basic components: multi-scale pose representation, motion encoding, and motion inference. In the following section, we \ywu{elaborate} the technical details of each component.  

\subsection{Multi-Scale Pose Representation}
We abstract a human pose step by step and obtain a series of poses from fine \ywu{scale} to coarse scale. We find that the motion in the coarser \ywu{scale} is more stable \ywu{than in the finer scale}, for which the cross-modal motion synthesis is easier. Specifically, we adopt four scales for human pose representation: initial scale $\boldsymbol{S}_{1}$, coarse scale $\boldsymbol{S}_{2}$, coarser scale $\boldsymbol{S}_{3}$, and coarsest scale $\boldsymbol{S}_{4}$. 

\begin{figure}[!t]
	\centering
	\includegraphics[width=0.47\textwidth]{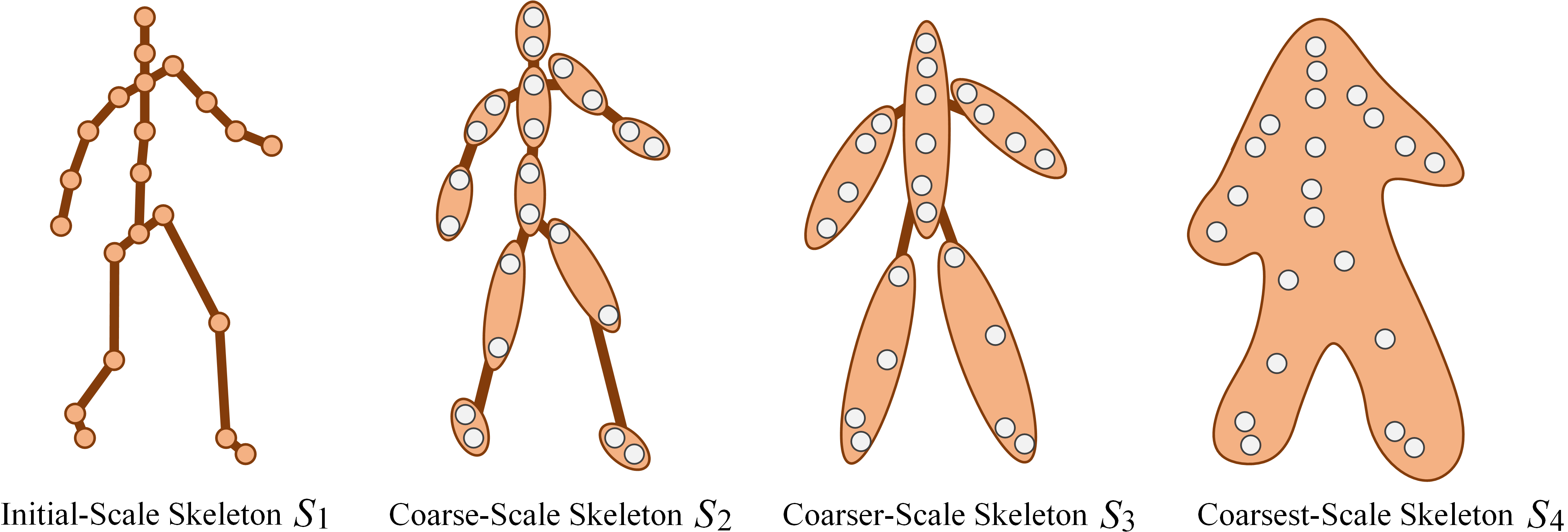}  
	\caption{Four body scales on HumanML3D dataset. In the initial scale $S_{1}$, each pose of HumanML3D skeleton contains 22 body-joint nodes. In $S_{2}$, $S_{3}$ and $S_{4}$, we consider 11, 5 and 1 body-part nodes, respectively.}
	\label{multi_scale}
\end{figure}

As shown in Figure \ref{multi_scale}, based on human body prior, we average the 3D position of spatially nearby $\boldsymbol{S}_{1}$ joints to \ywu{generate} the new position feature of $\boldsymbol{S}_{2}$, $\boldsymbol{S}_{3}$, and  $\boldsymbol{S}_{4}$. We further extrapolate the 3D rotation, velocities, and foot contact information of $\boldsymbol{S}_{2}$ and $\boldsymbol{S}_{3}$ based on their 3D joint positions and kinematic chains. \ywu{Every abstraction step follows the physical structure of human body and is made to be as interpretable as possible, so that the resulting nodes and overall poses can have meaningful correspondences with certain textual descriptions.} As an extreme body abstraction, the position information of $\boldsymbol{S}_{4}$ \ywu{encapsulates} the pose-insensitive trajectory cues of a motion sequence. 

\begin{figure*}[t]
	\centering
	\includegraphics[width=0.99\textwidth]{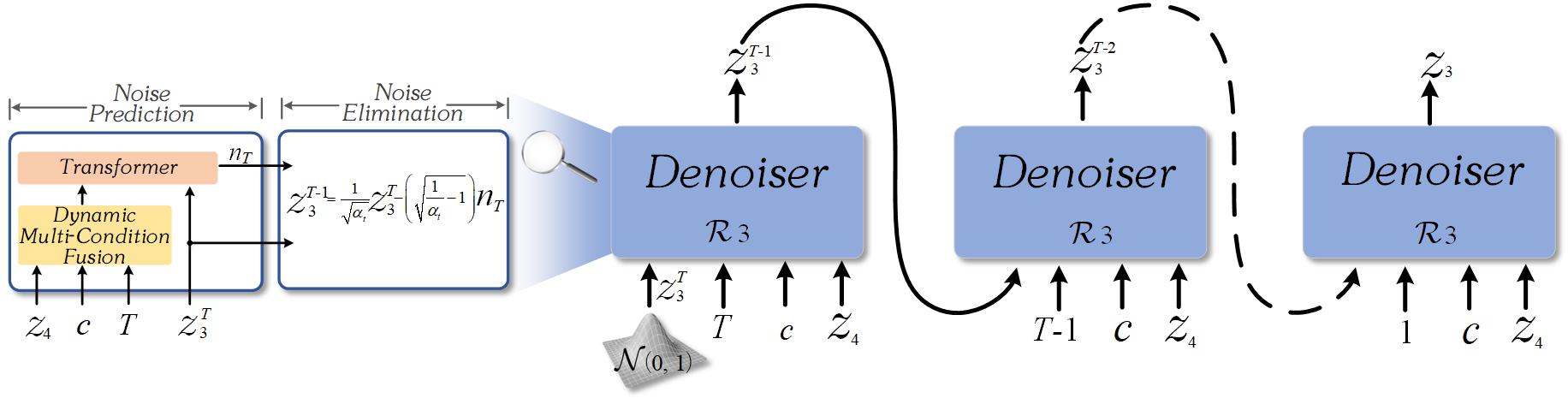}  
	\vspace{0.1in}
	\caption{Iterative Denoising Module. Taking denoiser $\mathcal{R}_{3}$ as an example, its denoising process can be factorized into two sequential stages: \textit{Noise Prediction} and \textit{Noise Elimination}. Given the initial textual condition embedding $c$ and synthesized coarse human motion embedding $z_{4}$, $\mathcal{R}_{3}$ recursively infers the latent motion embedding $z_{3}$ from a sampled Gaussian noise signal $z_{T}$ with $T$ Markov denoising steps.}
	\label{iterative_denoising}
\end{figure*}

\begin{figure}[t]
	\centering
	\includegraphics[width=0.49\textwidth]{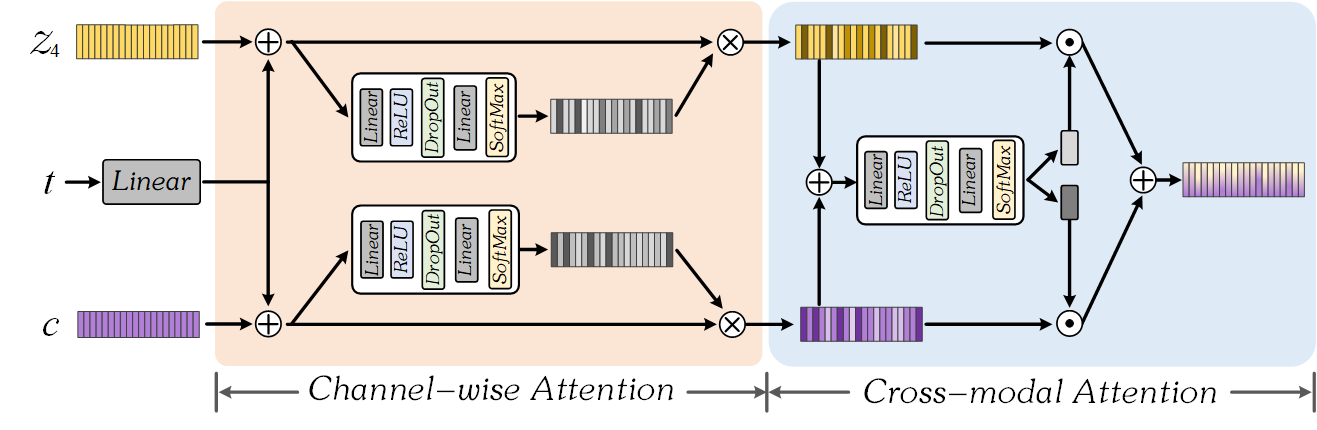}  
	\vspace{0.1in}
	\caption{Dynamic Multi-Condition Fusion Module. Taking the dynamic multi-condition fusion in $\mathcal{R}_{3}$ as an example, it adaptively infers the response score of $c$ and $z_{4}$ at $t$-th Markov denoising step. Based on the inferred channel-wise attention and cross-modal attention, $c$ and $z_{4}$ are integrated into a joint condition embedding effectively.}
	\label{dynamic_fusion}
\end{figure}

\subsection{Latent Motion Encoding}
We deploy a separate transformer-based Variational AutoEncoder $\mathcal{V}$ on each pose scale to encode it into a low-dimensional feature space $\mathcal{Z}$ and extract its $C_{e}$-dimensional motion embedding $z \in \mathbb{R}^{C_{e}}$. \revise{Specifically, at pose scale $\boldsymbol{S}_{i}$, $\mathcal{V}_{i}$ consists of a $L_{v}$-layer transformer-based encoder $\mathcal{E}_{i}$ and decoder $\mathcal{D}_{i}$ as $\mathcal{V}_{i}=\{\mathcal{E}_{i},\mathcal{D}_{i}\}$.} For simplicity, we take $\boldsymbol{S}_{1}$ as an example and introduce the technique in its latent motion representation learning. \ywu{The cases for other scales are similar.}
Given $\boldsymbol{S}_{1}$, its encoder $\mathcal{E}_{1}$ first embeds this raw motion sequence into a latent representation ${z}_{1}$. Similar to \cite{DBLP:conf/iccv/PetrovichBV21,DBLP:journals/corr/abs-2212-04048}, $\mathcal{E}_{1}$ uses the embedded distribution tokens as Gaussian distribution parameters $\mu_{1}$ and $\sigma_{1}$ of the latent embedding space to reparameterize $z_{1}$. Then, its decoder $\mathcal{D}_{1}$ is encouraged to reconstruct the 3D motion sequence $\widetilde{\boldsymbol{S}}_{1}$ from $z_{1}$. \revise{Finally, we train all VAEs  ($i.e.$, $\mathcal{V}_{1},\cdots,\mathcal{V}_{4}$) of four pose scales with two objectives ($i.e.$, $\mathcal{L}_{mr}$, $\mathcal{L}_{kl}$.) and optimize them end-to-end.} Specifically, $\mathcal{L}_{mr}$ defines a $l_2$-based motion reconstruction loss and focuses on learning an effective latent motion embedding $z$. To regularize the latent space $\mathcal{Z}_{i}$, $\mathcal{L}_{kl}$ computes a Kullback-Leibler distance between $q\left(z_{i}\mid S_{i}\right)=\mathcal{N}\left(z_{i}; \mathcal{E}_{\mu_{i}}, \mathcal{E}_{\sigma^2_{i}}\right)$ and a standard Gaussion distribution $\mathcal{N}\left(z_{i}; 0,1\right)$ at the $i$-th pose scale. Finally, the loss function of the VAE training stage of GUESS is defined as: 
	\begin{equation}
	\begin{aligned}
		\mathcal{L}_{\mathcal{V}} &= \lambda_{mr} \mathcal{L}_{mr} + \lambda_{kl} \mathcal{L}_{kl} \\
		&= \lambda_{mr} \sum_{i=1}^{4}\left\|\boldsymbol{S}_{i}-\mathcal{D}_{i}\left(\mathcal{E}_{i}\left(\boldsymbol{S}_{i}\right)\right)\right\|_2 \\
		&+ \lambda_{kl} \sum_{i=1}^{4}\textit{\textbf{KL}}\left(\mathcal{N}(\mu_{i}, \sigma^2_{i}) || \mathcal{N}(0, 1)\right), 
	\end{aligned}
	\label{eq.1}
\end{equation}
 where $\lambda_{mr}$ and $\lambda_{kl}$ represents the weight of $\mathcal{L}_{mr}$ and $\mathcal{L}_{kl}$, respectively.

\subsection{Cascaded Latent Diffusion}
\label{Cascaded Latent Diffusion}
Generally, denoising diffusion model \cite{DBLP:conf/icml/Sohl-DicksteinW15} learns the cross-modal mapping with the noising-denoising strategy. Specifically, in the training stage, our cascaded latent diffusion model first injects a random noise signal into four latent motion representations ($i.e.$, $z_{1}, \cdots, z_{4}$). Then, it iteratively denoises each motion embedding with the joint condition of given textual embedding and inferred coarser human motion embedding. In the test stage, based on the given CLIP-encoded text description, the cascaded latent diffusion model gradually enriches the human motion embedding by annealing a Gaussian noise signal step by step. In the following, we elaborate the technical details of this cascaded latent diffusion model.

\subsubsection{Forward Noising Diffusion}
Inspired by the stochastic diffusion process in Thermodynamics, the probabilistic diffusion of human motion representation is modeled as a Markov noising process inside the latent space as:
\begin{equation}
	q\left(z^{t}_{i} \mid z^{t-1}_{i}\right)=\mathcal{N}\left(\sqrt{\alpha^{t}} z^{t-1}_{i},\sqrt{1-\alpha^{t}} I\right),
	\label{eq.2}
\end{equation}
where $z^{t}_{i}$ denotes the latent motion representation of $i$-th pose scale at $t$-th Markov noising step, $\alpha^{t}$ is a hyper-parameters for the $t$-step sampling. Intuitively, Eq.\ref{eq.2} can be interpreted as sampling a noise from Gaussian distribution $\epsilon \sim \mathcal{N}(0,1)$ and then injecting it into $z^{t-1}_{i}$. Finally, a $T_{i}$-length Markov noising process gradually injects random noise into the latent motion representation $z_{i}$ and arrives at a noising sequence $\{z^{t}_{i}\}_{t=0}^{T_{i}}$. If $T_{i}$ is sufficiently large, $z_{i}^{T_{i}}$ will approximate a normal Gaussian noise signal.

\revise{To learn an effective text-to-motion probabilistic mapping, as shown in Figure \ref{pipeline}, we develop a cascaded latent denoising diffusion model that builds a transformer-based denoiser $\mathcal{R}_{i}$ on the $i$-th scale and iteratively anneals the noise of $\{z^{t}_{i}\}_{t=0}^{T_{i}}$ to reconstruct $z_{i}$. Firstly, $\mathcal{R}_{4}(z_{4}^{t},t,c)$ learns the conditional distribution $p_{4}(z_{4}|c)$ at the $t$-th denoising step as:
\begin{equation}
	z_{4}^{t-1}=\frac{1}{\sqrt{\alpha}_t} z_{4}^{t} - \sqrt{\frac{1}{\alpha_{t}}-1} \mathcal{R}_{4}(z_{4}^{t},t,c), \text{where} \; 1\!\leq\!t\!\leq\!T_{4}. 
	\label{eq.3}
\end{equation}
Iterating these diffusion-based $T_{4}$ Markov denoising steps, we reconstruct the coarsest motion representation $z_{4}$ from a noise sequence $\{z^{t}_{4}\}_{t=0}^{T_{4}}$ with the guidance of textual conditional embedding $c$.}
\subsubsection{Dynamic Multi-Condition Fusion}
\revise{Given the coarsest human motion embedding $z_{4}$ as an additional prompt, the denoisers following $\mathcal{R}_{4}$ learn the joint condition distribution $p_{i}(z_{i}|z_{i-1},c)$ and iteratively denoises each motion embedding with the joint condition of given textual embedding $c$ and inferred coarser human motion embedding $z_{i-1}$.} Notably, considering that $c$ and $z_{i-1}$ play different roles inside the joint generation condition, we thus propose a dynamic multi-condition fusion module to adaptively infer their conditional weights in each input sample and denoising step.

For simplicity, we take $\mathcal{R}_{3}$ as an example and introduce the technique in its Markov denoising process. \ywu{The cases for other denoisers are similar.} As shown in Figure \ref{iterative_denoising}, its $t$-th denoising process can be factorized into two sequential stages: \textit{Noise Prediction} and \textit{Noise Elimination}. Firstly, at the \textit{Noise Prediction} stage, $\mathcal{R}_{3}$ deploys a dynamic multi-condition module that takes $\{z_{4}, t, c\}$ as its input tuple and infers the channel-wise attention and cross-modal attention of $z_{4}$ and $c$ at $t$-th denoising step.

The overview of the proposed dynamic multi-condition fusion module is shown in Figure \ref{dynamic_fusion}. To be sensitive to the denoising stage, we first deploy a linear projection $\theta(\cdot)$ to map $t$ into a $C_{e}$-dimensional embedding and introduce it into $z_{4}$ and $c$ as: $\widetilde{z_{4}}=z_{4} + \theta(t)$, $\widetilde{c}=c+\theta(t)$. Then, given $\widetilde{z_{4}}$ and $\widetilde{c}$, we input them into two independent non-linear projection layers to infer the feature response scores of their $C_{e}$ channels as:
\begin{equation}
	{\widehat{z_4}}={\widetilde{z_4}} \otimes \operatorname{SoftMax}\left(\theta_z^2\left(\sigma\left(\theta_z^1\left({\widetilde{z_4}}\right)\right)\right)\right),
	\label{eq.4}
\end{equation}
\begin{equation}
	{\widehat{c}}={\widetilde{c}} \otimes \operatorname{SoftMax}\left(\theta_c^2\left(\sigma\left(\theta_c^1\left({\widetilde{c}}\right)\right)\right)\right), \;  \; \;
	\label{eq.5}
\end{equation}
where $\sigma$ is a ReLU non-linearity and $\otimes$ denotes channel-wise product. After that, the feature embedding of $\widetilde{z_4}$ and $\widetilde{c}$ are dynamically refined based on their channel-wise responses and updated as ${\widehat{z_4}}$ and ${\widehat{c}}$, respectively. Finally, we infer the cross-modal attention to dynamically balance the cooperative effects of ${\widehat{z_4}}$ and ${\widehat{c}}$ in their joint condition $j_{t}$ at $t$-th denoising step as:
\begin{equation}
\begin{aligned}
	j_{t} &= w_{z} {\widehat{z_4}} + w_{c} \widehat{c}, \\
	\textit{where} \; [w_{z},w_{c}] &= \operatorname{SoftMax} \left(\theta_j^2\left(\sigma\left(\theta_j^1\left({\widehat{z_4}}+{\widehat{c}}\right)\right)\right)\right)
\end{aligned} 
 	\label{eq.6}
\end{equation}
\subsubsection{Reverse Iterative Denoising}
Given the joint embedding $j_{t}$ of $c$ and $z_{4}$, $\mathcal{R}_{3}$ deploys a $L_{\mathcal{R}}$-layer transformer $\mathcal{T}_{3}$ to infer the noise signal injected at $t$-th Markov diffusion step and denoises $z_{3}^{t}$ as:
\begin{equation}
z^{t-1}_{3}=\mathcal{R}_{3}(z_{3}^{t},t,c,z_{4})=\frac{1}{\sqrt{\alpha}_t} z_{3}^{t} - \sqrt{\frac{1}{\alpha_{t}}-1} \mathcal{T}_{3}(j_{t},z_{3}^{t}). 
	\label{eq.7}
\end{equation}
Recurring the these denoising steps $T$ times, $\mathcal{R}_{3}$ reconstruct $z_{3}$ from $\{z_{3}^{t}\}^{T_{3}}_{t=0}$ iteratively. Similar to $\mathcal{R}_{3}$, $\mathcal{R}_{2}$ infers $z_{2}$ from the textual embedding $c$ and coarser human motion embedding $z_{3}$. Finally, based on $c$ and $z_{2}$, $\mathcal{R}_{1}$ further enriches the human motion embedding and obtains $z_{1}$. We train all these denoisers ($i.e.,$ $\mathcal{R}_{1},\cdots,\mathcal{R}_{4}$) end-to-end, and the training objective \ywu{for the motion inference part} is defined as the denoising loss at \ywu{the $t$-th Markov step:} 
\begin{equation}
	\begin{aligned}
		\mathcal{L}_{MI}  = &\ \mathbb{E} \left[\left\|\epsilon-\mathcal{R}_{4}\left(z_{4}^{t}, t,c\right)\right\|_2^2\right] \\  & +
		\sum_{i=1}^{3}
		\mathbb{E} \left[\left\|\epsilon-\mathcal{R}_{i}\left(z_{i}^{t}, t,c, z_{i+1} \right)\right\|_2^2\right] \ywu{.}
	\end{aligned}
	\label{eq.8}
\end{equation}

In the inference \ywu{phase}, based on the textual condition embedding $c$, the \ywu{trained} cascaded latent denoising diffusion model denoises a Gaussian noise signal to infer the motion embedding \ywu{$z_{4}$}\ywu{, and then progressively generates $z_{3}$, $z_{2}$, until finally getting to $z_{1}$}. \ywu{After that}, the decoder $D_{1}$ \ywu{projects $z_{1}$} to the 3D pose space \ywu{to synthesize the final} human motion \ywu{sequence} $\boldsymbol{S}$.

\begin{table*}[t]
	\def\arraystretch{1.1}
	
	\centering
	\scalebox{0.97}{
		
		\begin{tabular}{l c c c c c c c}
			\toprule
			\multirow{2}{*}{Methods}  & \multicolumn{3}{c}{R-Precision $\uparrow$} & \multirow{2}{*}{FID $\downarrow$} & \multirow{2}{*}{MM Dist $\downarrow$} & \multirow{2}{*}{Diversity $\uparrow$} & \multirow{2}{*}{MModality $\uparrow$}\\
			
			\cline{2-4}
			~ & Top-1 & Top-2 & Top-3 \\
			
			\midrule
			
			\textbf{Real motion} & \et{0.511}{.003} & \et{0.703}{.003} & \et{0.797}{.002} & \et{0.002}{.000} & \et{2.974}{.008} & \et{9.503}{.065} & -  \\
			\midrule
			Seq2Seq \cite{DBLP:conf/seq2seq} & \et{0.180}{.002} & \et{0.300}{.002} & \et{0.396}{.002} & \et{11.75}{.035} & \et{5.529}{.007} & \et{6.223}{.061}  & -  \\
			
			Language2Pose \cite{DBLP:conf/3dim/AhujaM19} & \et{0.246}{.002} & \et{0.387}{.002} & \et{0.486}{.002} & \et{11.02}{.046} & \et{5.296}{.008} & \et{7.676}{.058} & -  \\
			
			Text2Gesture \cite{DBLP:conf/vr/BhattacharyaRBG21} & \et{0.165}{.001} & \et{0.267}{.002} & \et{0.345}{.002} & \et{5.012}{.030} & \et{6.030}{.008} & \et{6.409}{.071} & -  \\
			
			Hier \cite{DBLP:conf/iccv/GhoshCOTS21} & \et{0.301}{.002} & \et{0.425}{.002} & \et{0.552}{.004} & \et{6.532}{.024} & \et{5.012}{.018} & \et{8.332}{.042} & -  \\
			
			MoCoGAN \cite{DBLP:conf/cvpr/Tulyakov0YK18} & \et{0.037}{.000} & \et{0.072}{.001} & \et{0.106}{.001} & \et{94.41}{.021} & \et{9.643}{.006} & \et{0.462}{.008} & \et{0.019}{.000}  \\
			
			Dance2Music \cite{DBLP:conf/nips/LeeY0WLYK19} & \et{0.033}{.000} & \et{0.065}{.001} & \et{0.097}{.001} & \et{66.98}{.016} & \et{8.116}{.006} & \et{0.725}{.011} & \et{0.043}{.001}  \\
			
			TM2T \cite{DBLP:conf/eccv/GuoZWC22} & \et{0.424}{.003} & \et{0.618}{.003} & \et{0.729}{.002} & \et{1.501}{.017} & \et{3.467}{.011} & \et{8.589}{.076} & \et{2.424}{.093}  \\
			
			T2M \cite{DBLP:conf/cvpr/GuoZZ0JL022} & \et{0.457}{.002} & \et{0.639}{.003} & \et{0.740}{.002} & \et{1.067}{.002} & \et{3.340}{.008} & \et{9.188}{.002} & \et{2.090}{.083}  \\
			MDM \cite{DBLP:journals/corr/abs-2209-14916} & \et{0.320}{.005} & \et{0.498}{.004} & \et{0.611}{.007} & \et{0.544}{.044} & \et{5.566}{.027} & \et{9.559}{.086} & \etr{2.799}{.072}   \\
			MLD \cite{DBLP:journals/corr/abs-2212-04048} & \et{0.481}{.003} & \et{0.673}{.003} & \et{0.772}{.002} & \et{0.473}{.013} & \et{3.196}{.010} & \etbb{9.724}{.082} & \et{2.413}{.079} \\
			T2M-GPT \cite{DBLP:journals/corr/abs-2301-06052} & \etbb{0.492}{.003} & \etbb{0.679}{.002} & \etbb{0.775}{.002} & \etbb{0.141}{.005} & \etbb{3.121}{.009} & \et{9.722}{.082} & \et{1.831}{.048}    
			\\
			\midrule  
			\textbf{GUESS} (Ours) & \etr{0.503}{.003} & \etr{0.688}{.002} & \etr{0.787}{.002} & \etr{0.109}{.007} & \etr{3.006}{.007} & \etr{9.826}{.104} & \etbb{2.430}{.100} \\
			\bottomrule
		\end{tabular}
	}
	\vspace{1.5mm}
	\caption{Comparison of text-to-motion synthesis on HumanML3D dataset. Following the common-used evaluation scheme, we repeat the evaluation 20 times and report the average with 95\% confidence interval. The best and second-best results are bolded and underlined, respectively.}
	\label{tab1}
\end{table*}

\begin{table*}[t]
	\def\arraystretch{1.1}
	
	\centering
	\scalebox{0.97}{
		
		\begin{tabular}{l c c c c c c c}
			\toprule
			\multirow{2}{*}{Methods}  & \multicolumn{3}{c}{R-Precision $\uparrow$} & \multirow{2}{*}{FID $\downarrow$} & \multirow{2}{*}{MM-Dist $\downarrow$} & \multirow{2}{*}{Diversity $\uparrow$} & \multirow{2}{*}{MModality $\uparrow$}\\
			
			\cline{2-4}
			~ & Top-1 & Top-2 & Top-3 \\
			
			\midrule
			\textbf{Real motion} & \et{0.424}{.005} & \et{0.649}{.006} & \et{0.779}{.006} & \et{0.031}{.004} & \et{2.788}{.012} & \et{11.08}{.097} & -  \\
			\midrule
			Seq2Seq \cite{DBLP:conf/seq2seq} & \et{0.103}{.003} & \et{0.178}{.005} & \et{0.241}{.006} & \et{24.86}{.348} & \et{7.960}{.031} & \et{6.744}{.106}  & -  \\
			
			Language2Pose \cite{DBLP:conf/3dim/AhujaM19} & \et{0.221}{.005} & \et{0.373}{.004} & \et{0.483}{.005} & \et{6.545}{.072} & \et{5.147}{.030} & \et{9.073}{.100} & -  \\
			
			Text2Gesture \cite{DBLP:conf/vr/BhattacharyaRBG21} & \et{0.156}{.004} & \et{0.255}{.004} & \et{0.338}{.005} & \et{12.12}{.183} & \et{6.964}{.029} & \et{9.334}{.079} & -  \\
			
			Hier \cite{DBLP:conf/iccv/GhoshCOTS21} & \et{0.255}{.006} & \et{0.432}{.007} & \et{0.531}{.007} & \et{5.203}{.107} & \et{4.986}{.027} & \et{9.563}{.072} & -  \\
			
			MoCoGAN \cite{DBLP:conf/cvpr/Tulyakov0YK18} & \et{0.022}{.002} & \et{0.042}{.003} & \et{0.063}{.003} & \et{82.69}{.242} & \et{10.47}{.012} & \et{3.091}{.043} & \et{0.250}{.009}  \\
			
			Dance2Music \cite{DBLP:conf/nips/LeeY0WLYK19} & \et{0.031}{.002} & \et{0.058}{.002} & \et{0.086}{.003} & \et{115.4}{.240} & \et{10.40}{.016} & \et{0.241}{.004} & \et{0.062}{.002}  \\
			TM2T \cite{DBLP:conf/eccv/GuoZWC22} & \et{0.280}{.005} & \et{0.463}{.006} & \et{0.587}{.005} & \et{3.599}{.153} & \et{4.591}{.026} & \et{9.473}{.117} & \etr{3.292}{.081}  \\
			T2M \cite{DBLP:conf/cvpr/GuoZZ0JL022} & \et{0.361}{.006} & \et{0.559}{.007} & \et{0.681}{.007} & \et{3.022}{.107} & \et{3.488}{.028} & \et{10.720}{.145} & \et{2.052}{.107}  \\
			MDM \cite{DBLP:journals/corr/abs-2209-14916} & \et{0.164}{.004} & \et{0.291}{.004} & \et{0.396}{.004} & \et{0.497}{.021} & \et{9.191}{.022} & \et{10.847}{.109} & \et{1.907}{.214}  \\
			MLD \cite{DBLP:journals/corr/abs-2212-04048} & \et{0.390}{.008} & \et{0.609}{.008} & \et{0.734}{.007} & \etbb{0.404}{.027} & \et{3.204}{.027} & \et{10.80}{.117} & \et{2.192}{.071} \\
			T2M-GPT \cite{DBLP:journals/corr/abs-2301-06052} & \etbb{0.416}{.006} & \etbb{0.627}{.006} & \etbb{0.745}{.006} & \et{0.514}{.029} & \etbb{3.007}{.023} & \etbb{10.921}{.108} & \et{1.570}{.039}   
			\\
			\midrule
			\textbf{GUESS} (Ours) & \etr{0.425}{.005} & \etr{0.632}{.007} & \etr{0.751}{.005} & \etr{0.371}{.020} & \etr{2.421}{.022} & \etr{10.933}{.110} & \etbb{2.732}{.084} \\
			
			\bottomrule
		\end{tabular}
	}
	\vspace{1.5mm}
	\caption{Comparison of text-to-motion synthesis on KIT-ML dataset. Following the common-used evaluation scheme, we repeat the evaluation 20 times and report the average with 95\% confidence interval. The best and second-best results are bolded and underlined, respectively.}
	\label{tab2}
\end{table*}

\begin{table*}[h]
	\def\arraystretch{1.1}
\centering
\scalebox{0.8}{
		\begin{tabular}{@{}lccccccccc@{}}
			\toprule
			\multirow{2}{*}{Methods} & \multicolumn{5}{c}{UESTC}                                               & \multicolumn{4}{c}{HumanAct12}                          \\ \cmidrule(lr){2-6} \cmidrule(l){7-10} 
			& $\text{FID}_{\text{train}}\downarrow $            & $\text{FID}_{\text{test}}\downarrow$             & ACC$\uparrow$ & Diversity$\uparrow$ & MModality$\uparrow$ &   \multicolumn{1}{c}{$\text{FID}_{\text{train}}\downarrow$} & ACC $\uparrow$& Diversity$\uparrow$ & MModality$\uparrow$ \\ 
			\toprule
			
			\textbf{Real motion} & \et{2.92}{.26} & \et{2.79}{.29} & \et{0.988}{.001} & \et{33.34}{.320} & \et{14.16}{.06} & \et{0.020}{.010} & \et{0.997}{.001} & \et{6.850}{.050} & \et{2.450}{.040} \\ \midrule
			ACTOR \cite{DBLP:conf/iccv/PetrovichBV21} & \et{20.5}{2.3} & \et{23.43}{2.20} & \et{0.911}{.003} & \et{31.96}{.33} & \et{14.52}{.09} & \et{0.120}{.000} & \et{0.955}{.008} & \et{6.840}{.030} & \et{2.530}{.020} \\
			INR \cite{DBLP:conf/eccv/CervantesSSS22} & \et{9.55}{.06} & \et{15.00}{.090} & \et{0.941}{.001} & \et{31.59}{.19} & \et{14.68}{.07} & \et{0.088}{.004} & \et{0.973}{.001} & \etr{6.881}{.048} & \et{2.569}{.040} \\
            \revise{T2M* \cite{DBLP:conf/cvpr/GuoZZ0JL022}} & \et{10.79}{1.21} & \et{13.40}{.090} & \et{0.944}{.002} & \et{32.03}{.14} & \et{14.41}{.05} & \et{0.081}{.003} & \et{0.978}{.001} & \et{6.843}{.037} & \et{2.534}{.045} \\
			MDM \cite{DBLP:journals/corr/abs-2209-14916} & \et{9.98}{1.33} & \et{12.81}{1.46} & \et{0.950}{.000} & \et{33.02}{.28} & \et{14.26}{.12} & \et{0.100}{.000} & \et{0.990}{.000} & \et{6.680}{.050} & \et{2.520}{.010} \\
			MLD \cite{DBLP:journals/corr/abs-2212-04048} & \et{12.89}{.109} & \et{15.79}{.079} & \et{0.954}{.001} & \et{33.52}{.14} & \et{13.57}{.06} & \et{0.077}{.004} & \et{0.964}{.002} & \et{6.831}{.050} & \etr{2.824}{.038} \\ 
            \revise{T2M-GPT* \cite{DBLP:journals/corr/abs-2301-06052}} & \etbb{8.92}{1.01} & \etbb{11.31}{1.24} & \etbb{0.961}{.001} & \etbb{33.55}{.24} & \etbb{14.71}{.14} & \etbb{0.064}{.000} & \etbb{0.991}{.000} & \et{6.677}{.053} & \et{2.594}{.021} \\
   \midrule
			GUESS (Ours) & \etr{8.01}{.089} & \etr{9.59}{.060} & \etr{0.966}{.001} & \etr{33.59}{.14} & \etr{14.89}{.05} & \etr{0.051}{.002} & \etr{0.992}{.001} & \etbb{6.844}{.050} & \etbb{2.621}{.008}
			\\ \bottomrule
		\end{tabular}%
	}
	\vspace{4pt}
	\caption{Comparison of action-to-motion synthesis on UESTC and HumanAct12 datasets. $\text{FID}_\text{train}$, $\text{FID}_\text{train}$ and Accuracy (ACC) reflect the fidelity of generated motions. Diversity and MModality for motion diversity within each action label. The best and second-best results are bolded and underlined, respectively. \revise{* denotes the performances of the re-training model based on its official source codes.}}
	\label{tab:action2motion}
\end{table*}
\section{Experiments}
\subsection{Datasets}
	Text-driven conditional human motion supports rich data inputs, including an English-based textual description sentence and an action-specific type word. Compared with word-level conditions, textual description inputs have fine-grained annotations and offer more delicate details of motions, making the conditional human motion synthesis task more challenging. Therefore, we collect the following four datasets to evaluate the performance of GUESS. Specifically, we evaluate the text-to-motion synthesis on the first two datasets and evaluate the action-to-motion synthesis on the others.
 
\textbf{HumanML3D} \cite{DBLP:conf/cvpr/GuoZZ0JL022} is a common-used and large-scale dataset for the text-driven human motion synthesis task. It contains 14,616 human motions and 44,970 English-based sequence-level descriptions. Its raw motion samples are collected from AMASS \cite{DBLP:conf/iccv/MahmoodGTPB19} and HumanAct12 \cite{DBLP:conf/mm/GuoZWZSDG020}. The initial human skeleton scale in HumanML3D has 22 body joints, following their definitions in SMPL \cite{DBLP:conf/iccv/MahmoodGTPB19}. HumanML3D performs a series of data normalization operations: down-sampling the motion to 20 FPS; cropping the motion to 10 seconds.  Their text descriptions' average and median lengths are 12 and 10, respectively. 

\textbf{KIT Motion-Language (KIT-ML)} \cite{DBLP:journals/bigdata/PlappertMA16} contains 3,911 motion sequences and 6,278 textual descriptions. Its raw motion samples are collected from KIT \cite{DBLP:conf/icar/ManderyTDVA15} and Mocap \cite{Mocap} datasets. The initial human skeleton scale of KIT-ML has 21 body joints. Each motion sequence is down-sampled into 12.5 FPS and described by 1$\sim$4 English sentences.  

\textbf{HumanAct12} \cite{DBLP:conf/mm/GuoZWZSDG020} contains 1,191 motion clips with hierarchical word-level action type annotations. The collected actions are daily indoor activities and have 34 different categories, including \textit{warm up} and \textit{lift dumbbel}, \textit{etc.} Each skeletal body pose contains 24 joints. 

\textbf{UESTC} \cite{DBLP:conf/mm/JiXYSSZ18} contains 72,709 human motion samples over 40 word-level action type annotations. These collected human motion clips are captured from 118 subjects and 8 different viewpoints. All these factors enable the collect human motion data to be realistic and diverse and make the text-to-motion synthesis task more challenging.

\subsection{Implementation Details}
We set 4 scales, which contain initial body joints, 11, 5, and 1 body components for both datasets (shown in Figure \ref{multi_scale}). In the motion embedding, all encoders $\mathcal{E}_1\sim\mathcal{E}_{4}$ and decoders $\mathcal{D}_1\sim\mathcal{D}_{4}$ are 9-layer transformers with 8 heads. The number of channels for all motion embeddings $z_1\sim z_{4}$ is 512. In the motion inference, we employ a pre-trained \textit{CLIP-ViT-L-14} \cite{CLIP} as the text encoder and freeze its parameters in our training. The channel of text embedding $c$ is 512. In the cascaded latent denoising diffusion, each denoiser $\mathcal{R}$ is a 8-layer transformer with 4 heads. The step number of the Markov noising process on each scale is 250 ($i.e.$,$T_{1}=T_{2}=T_{3}=T_{4}=250$). We leave the investigation on these default configurations of GUESS in the following experiment section. Finally, we implement GUESS with PyTorch 1.3 on two RTX-3090 GPUs. The parameters of GUESS are optimized with a two-stage training scheme. The epochs of VAE and diffusion training stages are 1k and 4k, respectively. Our mini-batch size is set to 128 during VAE training and 64 during diffusion training. AdamW optimizes all parameters with initial learning rate $10^{-4}$.

\begin{figure*}[t]
	\centering
	\includegraphics[width=0.99\textwidth]{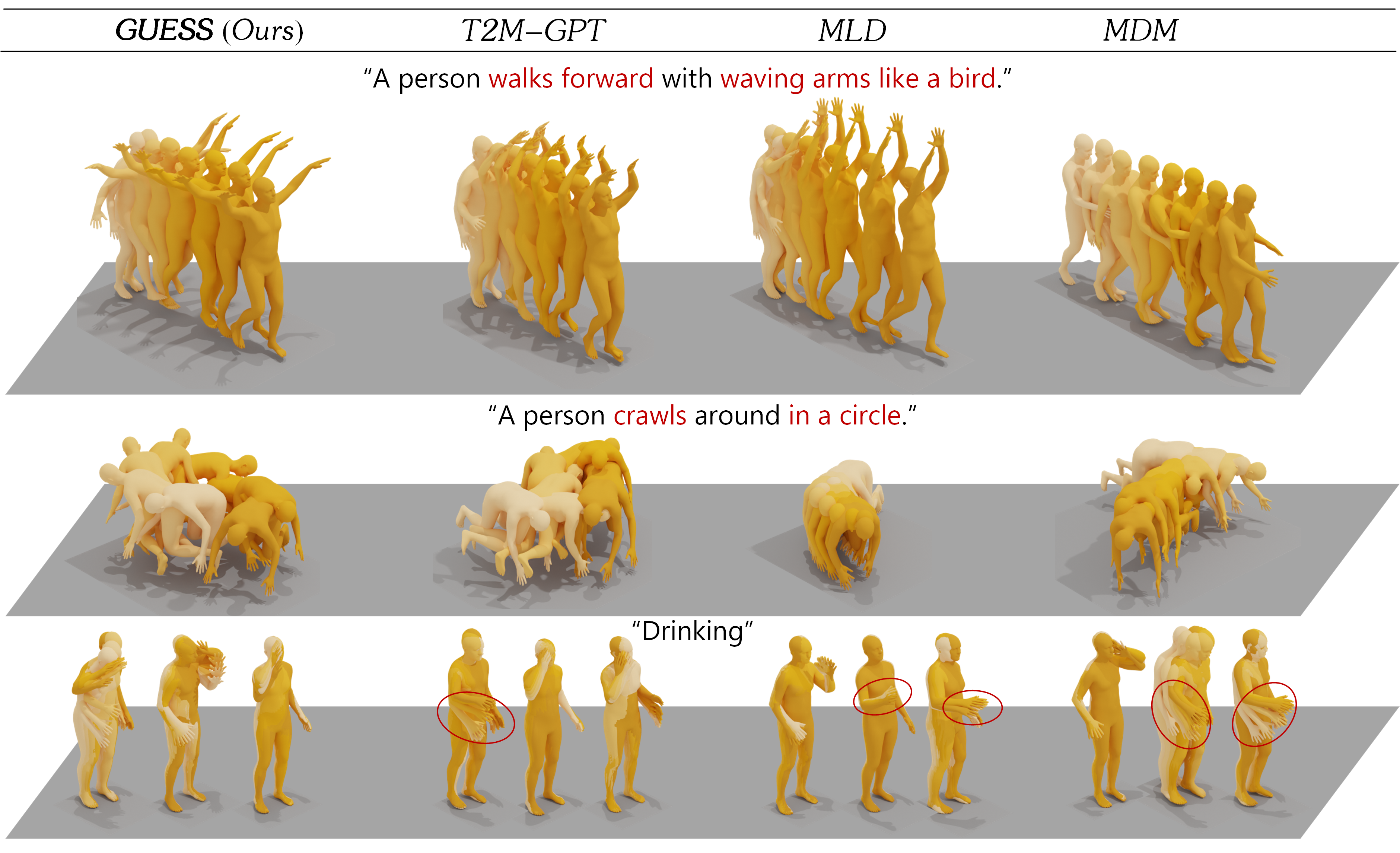}  
	\caption{Qualitative Comparison. We visualize generated human motion samples of GUESS and baseline methods on text-to-motion and action-to-motion evaluations. These qualitative evaluations indicate that GUESS generates realistic motions and significantly improves text-motion consistency.}
	\label{vis_compare}
\end{figure*}

\begin{figure*}[t]
	\centering
	\includegraphics[width=0.99\textwidth]{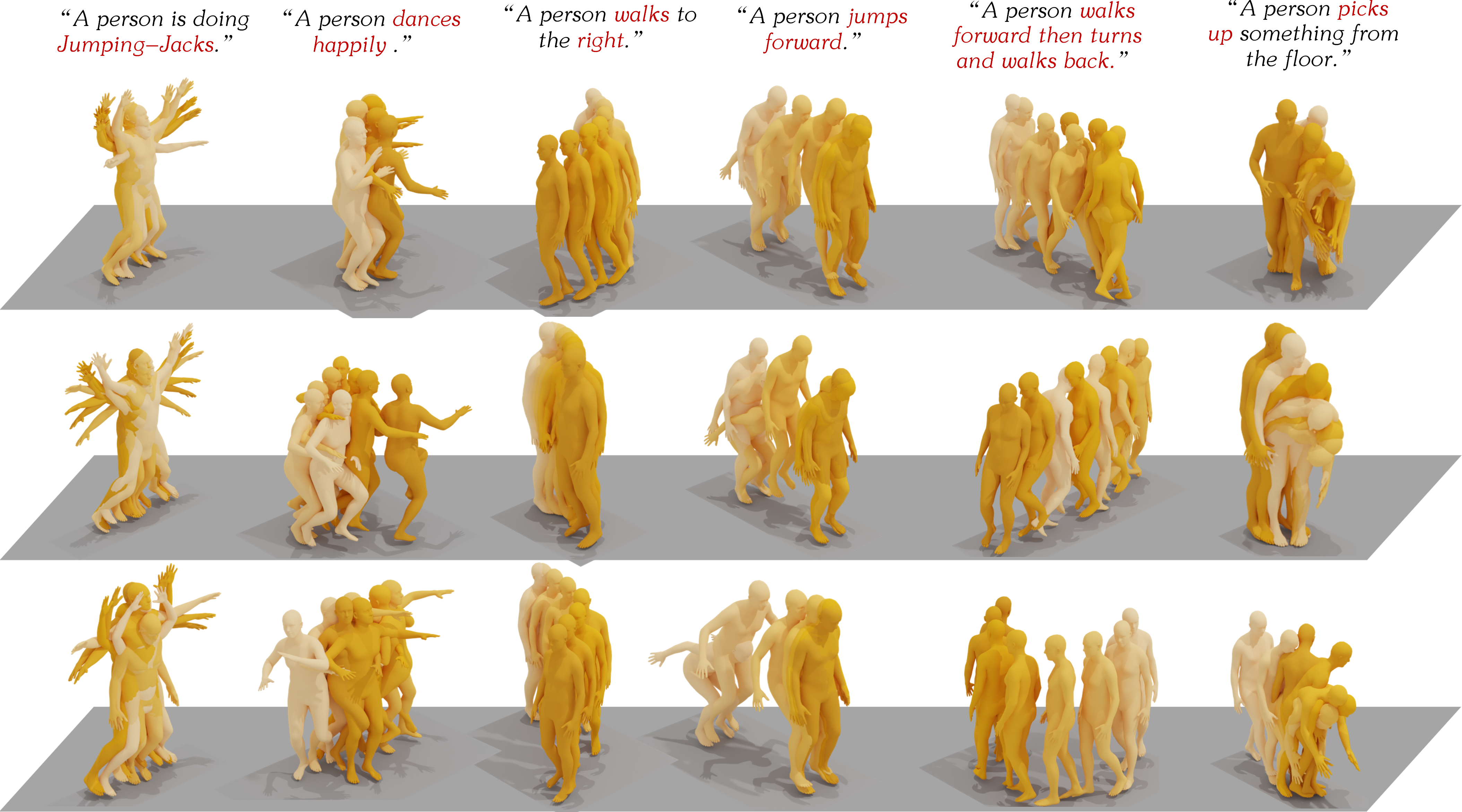}  
	\caption{Diverse synthesized motion samples. We visualize two human motion generation results for each textual description. These qualitative evaluations indicate that GUESS generates realistic and diverse motions.}
	\label{vis}
\end{figure*}

\subsection{Evaluation Metrics}
Following previous methods \cite{DBLP:conf/cvpr/GuoZZ0JL022,DBLP:journals/corr/abs-2212-04048}, we adopt five widely used quantitative metrics for text-to-motion synthesis methods to evaluate their performances. Besides, we further adopt Accuracy as one of the quantitative evaluation metrics for action-to-motion synthesis \cite{DBLP:journals/corr/abs-2212-04048}. These evaluation metrics analyze the performances of synthesis methods from the fidelity, text-motion consistency and diversity of their generated samples:
\begin{itemize}
	\item  \textit{R-Precision} reflects the text-motion matching accuracy in the retrieval. Given a motion sequence and 32 text descriptions (1 matched and 31 mismatched), we rank their Euclidean distances. The ground truth entry falling into the top-k candidates is treated as successful retrieval, otherwise it fails.
	\item \textit{Frechet Inception Distance} (FID) evaluates the feature distribution distance between the generated and real motions by feature extractor.
	\item \textit{Multi-Modal Distance} (MM Dist) is computed as the average Euclidean distance between the motion feature of each generated motion and the text feature of its description pair in the test set.
	\item \textit{Diversity} (DIV) is defined as the variance of motion feature vectors of the generated motions across all text descriptions, reflecting the diversity of synthesized motion from a set of various descriptions.
	\item \textit{Multi-modality} (MModality) measures how much the generated motions diversify within each text description, reflecting the diversity of synthesized motion from a specific description.
	\item \textit{Accuracy} (ACC) is computed as the average action recognition performance with generated motions, reflecting the fidelity of synthesized action-to-motion samples with given action types.   
\end{itemize}

\subsection{Evaluations on Text-to-motion}
In this section, we evaluate the text-to-motion synthesis performance of GUESS with quantitative and qualitative analyses on HumanML3D and KIT-ML datasets. These comparisons comprehensively evaluate synthesis methods from their performances on fidelity, text-motion consistency, and diversity.

\subsubsection{Quantitative Comparison}
In this section, we compare the quantitative performances of GUESS and previous works. Specifically, as shown in Table \ref{tab1}, GUESS significantly outperforms state-of-the-art methods on HumanML3D and achieves 23\% gains on FID performance, significantly improving the fidelity of generated human motions. The improvements in R-Precision verify that the human motions synthesized by GUESS have better text-motion consistency. Besides, better MM Dist and Diversity performances indicate that GUESS generates diverse human motion samples from the same textual input. Furthermore, Table \ref{tab2} verifies that GUESS also shows consistent superiority on the KIT-ML dataset. Compared with the state-of-the-art method T2M-GPT \cite{DBLP:journals/corr/abs-2212-04048}, GUESS outperforms it on all evaluation metrics, verifying our better performances on realistic, accurate and diverse text-to-motion generation. These quantitative performances on HumanML3D and KIT-ML datasets indicate that GUESS develops a strong baseline on these challenging text-driven human motion synthesis benchmarks.

\subsubsection{Qualitative Comparison}
In this section, we evaluate the performance of GUESS and baseline methods with qualitative comparisons. As described in Section \ref{method}, GUESS progressively enriches motion details on multiple granularity levels according to the given textual descriptions. Therefore, as verified in Figure \ref{vis_compare}, given the same textual condition input, the human motion samples generated by GUESS have more motion details and better text-motion consistency, significantly outperforming other baseline methods. Furthermore, as shown in the bottom of Figure \ref{vis_compare}, we also generate three motion samples for each given action label. Human action samples synthesized based on the same action category have different action details while conforming to the same action semantics.
\begin{figure}[t]
	\centering
	\includegraphics[width=0.49\textwidth]{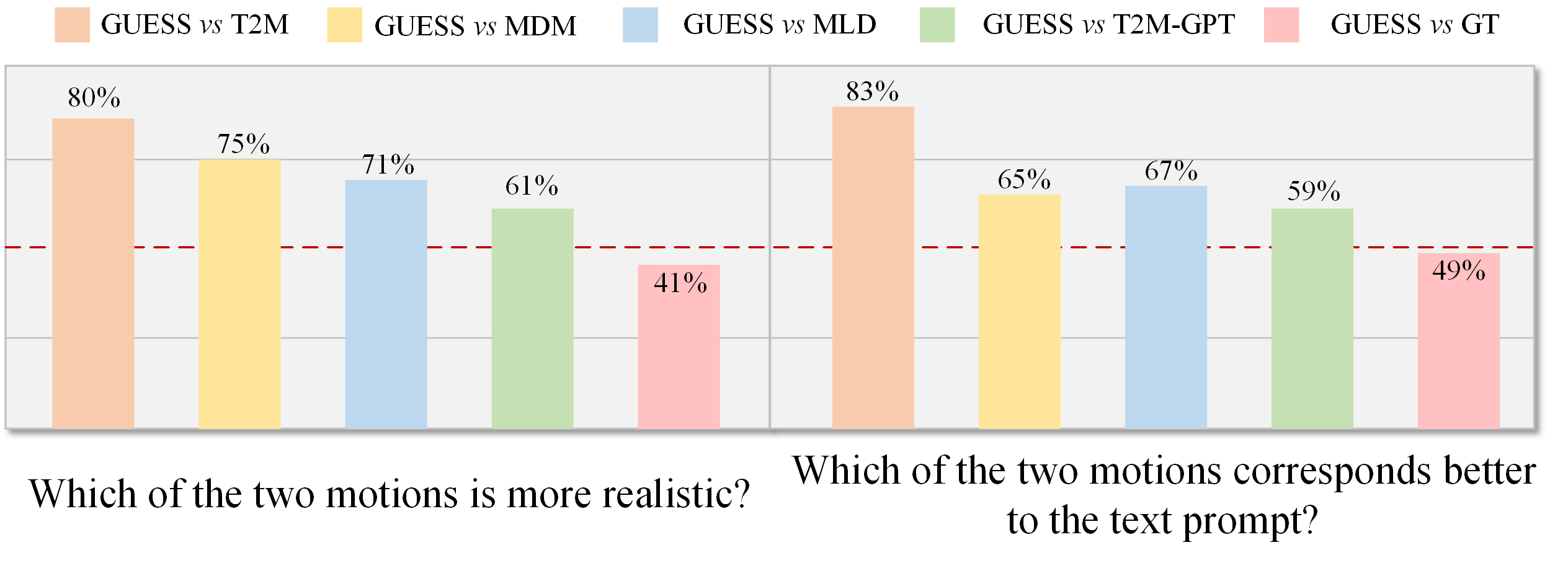}  
	\caption{User Study. Each bar indicates the preference rate of GUESS over other methods. The red line indicates the 50\%.}
	\label{user_study}
\end{figure}
\subsubsection{User Study}
\revise{In this section, we evaluate text-to-motion generation performances with perceptual user studies. Specifically, as shown in Fig. \ref{user_study}, we adopt a force-choice paradigm that asks “Which of the two motions is more realistic?” and “Which of the two motions corresponds better to the text prompt?”. The provided human motion samples are generated from 30 text condition inputs randomly selected from the test set of HumanML3D dataset. Then, we invite 20 subjects and provide five comparison pairs: ours vs T2M, ours vs MDM, ours vs MLD, ours vs T2M-GPT, and ours vs real motions from the dataset. As verified in Fig. \ref{user_study}, our method significantly outperforms the other state-of-the-art methods on motion realism and text-motion consistency and even competitive to the real motions.}

\subsection{Evaluation on Action-to-motion}
The action-to-motion synthesis task inputs a given action label and generates corresponding human motion samples. Here, we evaluate the action-to-motion performance of GUESS by comparing it with four baseline methods. Specifically, ACTOR \cite{DBLP:conf/iccv/PetrovichBV21} and INR \cite{DBLP:conf/eccv/CervantesSSS22} are transformer-based VAE models for the action-conditioned task. MDM \cite{DBLP:journals/corr/abs-2209-14916} and MLD \cite{DBLP:journals/corr/abs-2212-04048} adopt the diffusion-based generative framework and are state-of-the-art methods for text-driven human motion synthesis. As shown in Table \ref{tab:action2motion}, GUESS outperforms these competitors on all evaluation metrics of UESTC and HumanAct12 datasets, significantly improving realistic and diverse action-conditional human motion synthesis. These results verify that the proposed gradually enriching synthesis is a generic and powerful cross-modal generation strategy for both action-to-motion and text-to-motion tasks.

\subsection{Synthesized Sample Visualization}
In this section, we visualize more synthesized samples to evaluate the performance of GUESS, in terms of verisimilitude, diversity, and text-motion consistency. Specifically, we take five different textual condition inputs as examples and visualize two human motion samples generated from a same text condition. As shown in Figure \ref{vis}, we can see that GUESS can generate multiple plausible motions corresponding to the same text, performing diverse human motion generation. For example, GUESS generates diverse human walking actions with different motion speeds, paths, and styles. These visualization superiorities indicate that GUESS explores the freedom derived from the ambiguity behind linguistic descriptions while improving generation quality. Please refer to the supplemental demo video for more visualizations.
\begin{table}[t]
	\def\arraystretch{1.2}
	\resizebox{0.49\textwidth}{!}{
		\begin{tabular}{ccccc|cccc}
			\toprule
			\multicolumn{5}{c|}{Body Pose Scales} & \multirow{2}{*}{\begin{tabular}[c]{@{}c@{}}R-Precision \\ Top-1\end{tabular} $\uparrow$} & \multirow{2}{*}{FID $\downarrow$} & \multirow{2}{*}{Diversity $\uparrow$} & \multirow{2}{*}{Time $\downarrow$} \\
			\cline{1-5}
			$S_{1}$    & $S_{2}$    & $S_{3}$    & $S_{3}^{+}$    & $S_{4}$    &                                                                      &                      &                            &                           \\ \hline
			\checkmark  &            &            &            &            &  $0.475$ & $0.485$ & $9.703$  & $0.3$ \\
			\checkmark  & \checkmark &            &            &            &  $0.476$ & $0.394$ & $9.739$  & $0.5$ \\
			\checkmark  &            & \checkmark &            &            &  $0.476$ & $0.382$ & $9.742$  & $0.5$ \\
			\checkmark  &            &            & \checkmark &            &  $0.478$ & $0.345$ & $9.749$  & $0.5$ \\
			\checkmark  &            &            &            & \checkmark &  $0.479$ & $0.288$ & $9.756$  & $0.5$ \\
			\checkmark  &  \checkmark& \checkmark &            &            &  $0.486$ & $0.215$ & $9.760$  & $0.7$ \\
			\checkmark  &  \checkmark& \checkmark & \checkmark &            &  $0.490$ & $0.197$ & $9.763$  & $1.3$ \\
			\checkmark  &  \checkmark& \checkmark &            & \checkmark &  $\boldsymbol{0.503}$ & $0.109$ & $9.826$  & $1.3$ \\
			\checkmark  &  \checkmark& \checkmark & \checkmark &  \checkmark&  $\boldsymbol{0.503}$ & $\boldsymbol{0.108}$ & $\boldsymbol{9.827}$  & $1.9$ \\
			\bottomrule
		\end{tabular}
	}
	\vspace{1.5mm}
	\caption{The performance comparison between different pose scale configurations. The time performance we reported is the average inference time (second) of each sentence.}
	\label{tab3}
\end{table}

\subsection{Component Studies}
In this section, we analyze the individual components and investigate their configurations in the final GUESS architecture. Unless stated, the reported performances are Top-1 R-Precision, FID, and Diversity on HumanML3D dataset.

\subsubsection{Effect of Multiple Pose Scales} The intention that motivates us to tune the number of pose scales is twofold: (1) verify the effectiveness of the progressive generation scheme on multiple pose scales; (2) investigate the optimal number of multiple scales. Specifically, besides the four pose scales in our model, we further introduce an additional scales: $S_{3}^{+}$, which represents a body as 2 parts: upper body and lower body. Notably, Table \ref{tab3} verifies that using two pose scales ($e.g.$, $S_{1},S_{2}$ or $S_{1},S_{3}$) is significant better than using only initial $S_{1}$. Although Table \ref{tab3} reports that the combination of five pose scales achieves the best FID performance, considering the computational time cost, its performance improvements are limited. Therefore, we choose the combination of four pose scales ($S_{1},S_{2},S_{3}$ and $S_{4}$) as our final model configuration.  

\begin{figure}[t]
	\centering
	\includegraphics[width=0.48\textwidth]{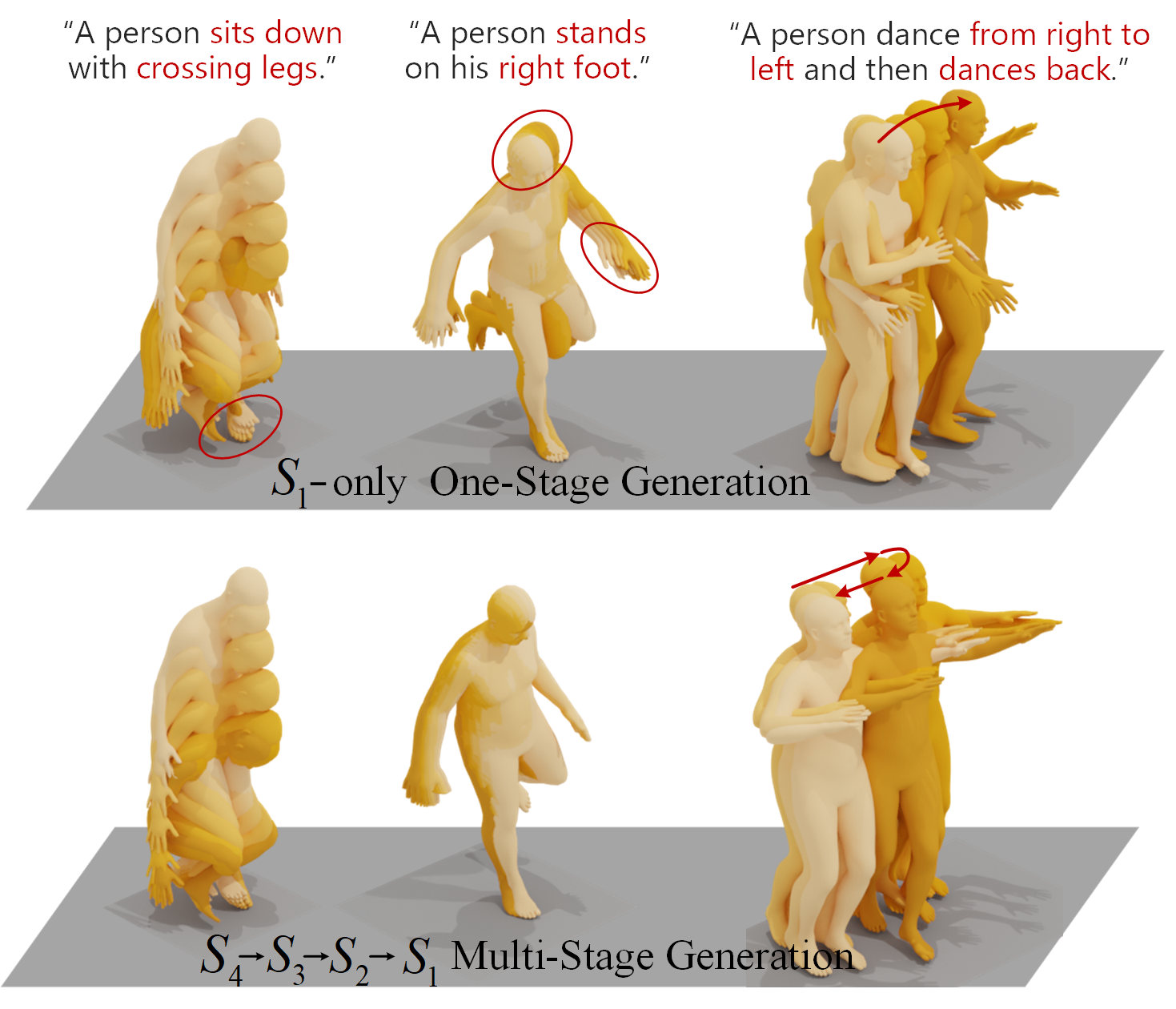}  
	\caption{Visual comparison between $S_{1}$-only one-stage generation and cascaded four-stage generation. As shown in red circles, the synthesized results of one-stage generation suffer from the body-joint jittering problem. In contrast, the proposed cascaded progressive generation scheme injects the textual description and coarse motion cues into the cross-modal generation, introducing richer condition information. Red arrows indicate body motion trajectories.}
	\vspace{0.15in}
	\label{vis_2}
\end{figure}

Furthermore, we also analyze the effectiveness of the cascaded multi-scale generation scheme from its visual performances. As shown in Figure \ref{vis_2}, we visualize the human pose synthesis results of two generation schemes: (1) $S_{1}$-only one-stage generation; (2) Cascaded multi-stage generation on $S_{1},S_{2},S_{3}$, and $S_{4}$. Analyzing the visual comparisons shown in Figure \ref{vis_2}, we have two core observations: (I) progressive multi-scale generation strategy introduces richer cooperative conditions of coarse body motion and textual description into final synthesized results, significantly alleviating the body-joint jittering problem. (II) progressive multi-scale generation strategy enforces stronger conditions on the body motion trajectory to make it consistent with the textual description, improving the generation quality.   

\begin{table}[t]
	\def\arraystretch{1.35}
	\resizebox{0.49\textwidth}{!}{
		\begin{tabular}{c|c|c|ccc|cc}
			\toprule\multicolumn{1}{c|}{\multirow{2}{*}{$\mathcal{E}$ \quad $\mathcal{D}$}} & \multicolumn{1}{c|}{\multirow{2}{*}{Layers}} & \multicolumn{1}{c|}{\multirow{2}{*}{Heads}} & \multicolumn{3}{c|}{VAE Reconstruction} & \multicolumn{2}{c}{Diffusion Synthesis} \\ \cline{4-8} 
			\multicolumn{1}{c|}{}                   & \multicolumn{1}{c|}{}                                  & \multicolumn{1}{c|}{}                                 & MPJPE $\downarrow$  & PAMPJPE $\downarrow$     & \multicolumn{1}{c|}{ACCL $\downarrow$}    & FID $\downarrow$   & DIV $\uparrow$          \\ \hline
			$\times$ & --- & --- & ---  &  ---  &  ---  &  $0.578$  & $9.369$ \\ \hline
			$\checkmark$ & $5$ & $4$ & $22.6$  &  $14.9$  &  $5.7$  &  $0.339$  & $9.559$ \\
			$\checkmark$ & $7$ & $4$ & $19.1$  &  $11.8$  &  $5.5$  & $0.278$  & $9.613$ \\
			$\checkmark$ & $9$ & $4$ & $15.7$  & $10.8$  & $5.4$  &  $0.215$  & $9.765$ \\
			$\checkmark$ & $9$ & $8$ & $\boldsymbol{13.9}$  & $\boldsymbol{9.2}$  & $\boldsymbol{5.3}$  &  $\boldsymbol{0.109}$  & $\boldsymbol{9.826}$ \\
			$\checkmark$ & $12$ & $12$ & $15.1$  & $9.9$  & $5.5$  &  $0.179$  & $9.801$ \\
			\bottomrule
		\end{tabular}
	}
	\vspace{1.5mm}
	\caption{The evaluation of our VAE module. Following \cite{DBLP:journals/corr/abs-2212-04048}, MPJPE, PAMPJPE, and ACCL are reported as motion reconstruction evaluation metrics.}
	\label{tab4}
\end{table}

\begin{figure}[t]
	\centering
	\includegraphics[width=0.48\textwidth]{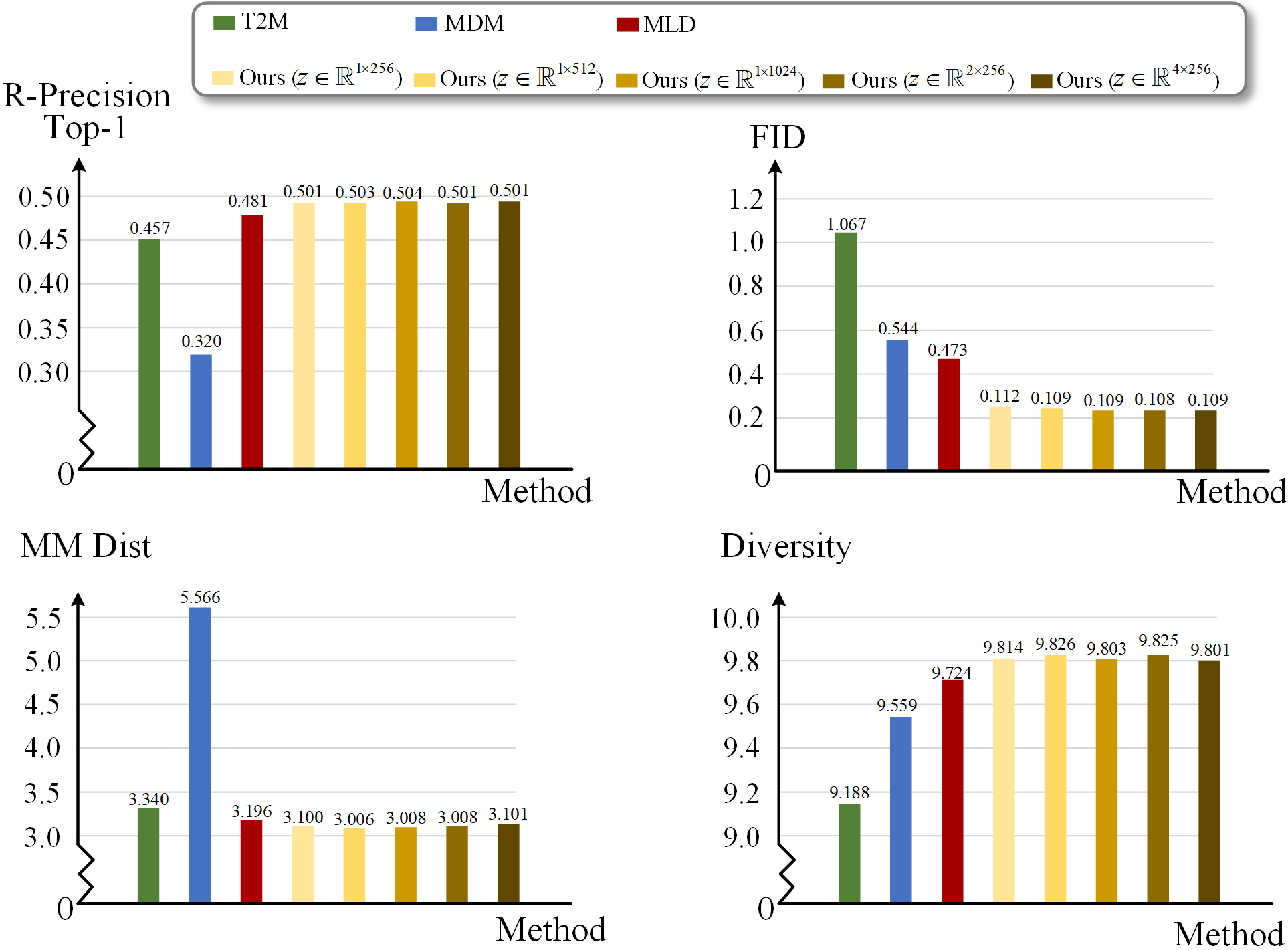}  
	\caption{The performance comparison between different shape configurations of latent motion embedding $z$. The performances of T2M \cite{DBLP:conf/cvpr/GuoZZ0JL022}, MDM \cite{DBLP:journals/corr/abs-2209-14916}, and MLD \cite{DBLP:journals/corr/abs-2212-04048} are introduced to analyze our performance gains.}
	\vspace{0.15in}
	\label{z_shape}
\end{figure}

\subsubsection{Effect of Motion Embedding} The intention that motivates us to tune the configuration of motion embedding is twofold: (1) investigate the effectiveness of latent motion embedding in the diffusion model; (2) choose the optimal number of layers and heads for transformer-based encoders $\mathcal{E}$ and decoders $\mathcal{D}$; (3) choose the optimal number of feature channels for latent motion embeddings $z$. Firstly, Table \ref{tab4} indicates that, without VAE, directly modelling the joint distribution over the raw motion sequences tends to hurt the generation quality. This observation is also supported by previous methods \cite{DBLP:journals/corr/abs-2212-04048,DBLP:conf/cvpr/RombachBLEO22}. Therefore, instead of using a diffusion model to establish the connections between the raw motion sequences and the conditional inputs, GUESS learns a probabilistic mapping from the conditions to the representative motion latent embeddings. In addition, we further tune the number of heads and layers of encoders and decoders to explore their optimal configurations. As reported in Table \ref{tab4}, GUESS achieves the best performance with 9 layers and 8 heads. An oversized VAE model tends to hurt motion reconstruction and synthesis. Besides, we further provide $5$ dimension configurations for the latent motion embedding $z$ and report their synthesis performances. To visually reflect the performance changes, we further introduce T2M \cite{DBLP:conf/cvpr/GuoZZ0JL022}, MDM \cite{DBLP:journals/corr/abs-2209-14916}, and MLD \cite{DBLP:journals/corr/abs-2212-04048} into our comparisons. As shown in Fig.\ref{z_shape}, GUESS is insensitive to the shape configuration of $z$, indicating that the generic progressive multi-stage generation strategy is the main reason for the observed performance gains.

\begin{table}[t]
	\def\arraystretch{1.35}
	\resizebox{0.49\textwidth}{!}{
		\begin{tabular}{c|c|c|c|ccc}
			\toprule
			\multirow{2}{*}{Denoiser} & \multicolumn{2}{c|}{Denoiser's input}                                            & \multirow{2}{*}{Layers} & \multirow{2}{*}{\begin{tabular}[c]{@{}c@{}}Top-1\\ R-Precision\end{tabular} $\uparrow$} & \multirow{2}{*}{FID $\downarrow$} & \multirow{2}{*}{DIV $\uparrow$} \\ \cline{2-3}
			& \multicolumn{1}{c|}{Noised Motion} & Conditions &     &                       &      \\ \hline
			\multirow{2}{*}{$\mathcal{R}_{4}$} & \multirow{2}{*}{ $z_{4}^{T_{4}}$} & \multirow{2}{*}{$c$} & 4 &  $0.491$ & $0.166$  & $9.744$ \\
			&   &  & 8 &  $\boldsymbol{0.503}$ & $\boldsymbol{0.109}$  & $\boldsymbol{9.826}$ \\ \hline
			\multirow{3}{*}{$\mathcal{R}_{3}$} & \multirow{3}{*}{ $z_{3}^{T_{3}}$} & \revise{$c$} & \revise{4} &  \revise{$0.493$} & \revise{$0.153$}  & \revise{$9.701$} \\ \cline{3-7}
			&   & \multirow{2}{*}{[$c$, $z_{4}$]} & 4 &  $0.497$ & $0.121$  & $9.802$ \\ 
			&   &  & 8 &  $\boldsymbol{0.503}$ & $\boldsymbol{0.109}$  & $9.826$ \\ \hline
			\multirow{4}{*}{$\mathcal{R}_{2}$} & \multirow{4}{*}{ $z_{2}^{T_{2}}$} & \revise{$c$} &\revise{4} &  \revise{$0.494$} & \revise{$0.139$}  & \revise{$9.769$} \\ \cline{3-7}
			&   & \multirow{2}{*}{[$c$, $z_{3}$]} & 4 &  $0.494$ & $0.139$  & $9.769$ \\
			&   &  & 8 &  $\boldsymbol{0.503}$ & $\boldsymbol{0.109}$  & $9.826$ \\ \cline{3-7}
			&   & \revise{[$c$, $z_{3}$, $z_{4}$] } & \revise{4} &  \revise{$0.495$} & \revise{$0.137$}  & \revise{$9.772$} \\ \hline
			\multirow{5}{*}{$\mathcal{R}_{1}$} & \multirow{5}{*}{$z_{1}^{T_{1}}$} & $c$ & 4 &  $0.485$ & $0.266$  & $\boldsymbol{9.975}$ \\ \cline{3-7}
			&  & \multirow{2}{*}{ [$c$, $z_{2}$]}  & 4 &  $0.495$ & $0.136$  & $9.845$ \\ 
			&  &  & 8 &  $\boldsymbol{0.503}$ & $\boldsymbol{0.109}$  & $\boldsymbol{9.826}$ \\ \cline{3-7}
			&  & \revise{[$c$, $z_{2}, z_{3}$]}  & \revise{4} &  \revise{$0.498$} & \revise{$0.123$}  & \revise{$9.662$} \\ \cline{3-7}
			&  & \revise{[$c$, $z_{2}, z_{3}, z_{4}$]}  & \revise{4} & \revise{$0.500$} & \revise{$0.120$}  & \revise{$9.515$} \\ 
			\bottomrule
			
		\end{tabular}
	}
	\vspace{1.5mm}
	\caption{The evaluation of different condition inputs and the number of layers in the cascaded latent diffusion model.}
	\label{tab5}
\end{table}

\subsubsection{Effect of Cascaded Latent Diffusion}
As described in Section \ref{Cascaded Latent Diffusion}, GUESS deploys a denoiser $\mathcal{R}$ on each scale and builds a cascaded latent diffusion model. In this section, we verify the effectiveness of the proposed cascaded latent diffusion model and explore its optimal configurations. \revise{Specifically, as shown in Table \ref{tab5}, we analyze GUESS's generation performances under two configuration setups: (1) tuning the number of layers of all denoisers ($i.e.$, $\mathcal{R}_{1} \sim \mathcal{R}_{4}$) from 4 to 8; (2) tuning the cascading strategy among denoisers from sequential connection to dense connection. Analyzing the results shown in Table \ref{tab5}, our core observations are summarized into the following: (1) a 8-layer transformer is the best choice for each denoiser $\mathcal{R}_{i}$; (2) introducing cooperative condition of textual codition embedding and coarse motion embedding into the denoising process brings clear performance gains; (3) GUESS is insensitive to specific cascade strategies, indicating that the generic progressive generation scheme is the main reason for the observed improvements. Thus, for simplicity, we finally choose sequential cascading connections between four denoisers.}

\begin{table}[t]
	\def\arraystretch{1.35}
	\resizebox{0.49\textwidth}{!}{
\begin{tabular}{ccl|ccc}
	\toprule
	\multicolumn{3}{c|}{\begin{tabular}[c]{@{}c@{}}Dynamic Multi-Condition Fusion\\ Configuration\end{tabular}} & \multirow{2}{*}{\begin{tabular}[c]{@{}c@{}}R-Precision\\ Top-1\end{tabular} $\uparrow$}& \multirow{2}{*}{FID $\downarrow$} & \multirow{2}{*}{Diversity $\uparrow$} \\ \cline{1-3}
	Channel-wise Attention                            & \multicolumn{2}{c|}{Cross-Modal Attention}              &                                                                              &                      &                            \\ \hline
	$\times$   & \multicolumn{2}{c|}{$\times$}  & 0.489          & 0.162               & 0.961                          \\
	$\times$   & \multicolumn{2}{c|}{$\checkmark$} & 0.493       & 0.145                    & 0.972 \\
	$\checkmark$   & \multicolumn{2}{c|}{$\times$} & 0.497       & 0.137                    & 0.970 \\
	$\checkmark$   & \multicolumn{2}{c|}{$\checkmark$} & $\boldsymbol{0.503}$ & $\boldsymbol{0.109}$ & $\boldsymbol{9.826}$ \\ \bottomrule                           
\end{tabular}
	}
	\vspace{1.5mm}
	\caption{The performance comparison between different configurations of dynamic multi-condition fusion module.}
	\label{tab_dmcf}
\end{table}

\begin{figure}[t]
	\centering
	\includegraphics[width=0.48\textwidth]{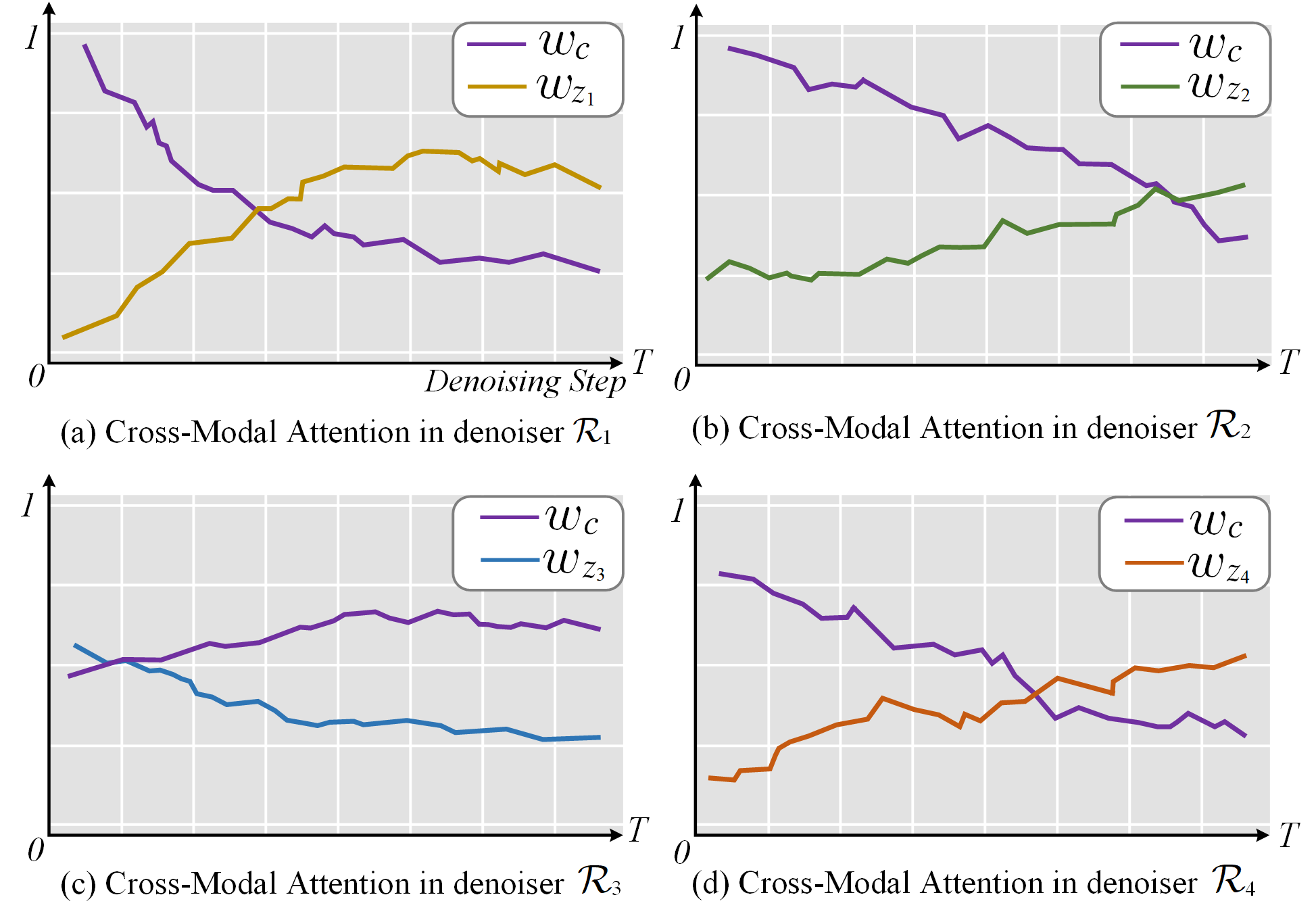}  
	\caption{The cross-modal attention weights of text embedding and motion embedding in different denoisers and denoising steps. Dynamic multi-condition fusion module adaptively infers $w_{z_{i}}$ and $w_{c}$ in denoiser $\mathcal{R}_{i}$ and its $T$ denoising steps.}
	\vspace{0.15in}
	\label{alpha}
\end{figure}

\subsubsection{Effect of Dynamic Multi-Condition Fusion} In this section, we analyze the effectiveness of the proposed dynamic multi-condition fusion module. Firstly, we investigate the effect of channel-wise attention and cross-modal attention on the final generation performance. As shown in Table \ref{tab_dmcf}, we find that integrating the dynamic multi-condition fusion module into the GUESS brings clear performance gains on Top-1 R-Precision and FID. Then, we further report the cross-modal attention responses ($i.e.$, $w_{z}$ and $w_c$) in four denoisers and $T$ denoising steps. Analyzing the results shown in Figure \ref{alpha}, we observe that the attention weights of text condition embedding and coarse motion embedding are dynamically tuned over different denoising stages, adaptively balancing their effects on conditional motion generation. All these results verify the effectiveness of the dynamic multi-condition fusion in text-driven human motion synthesis.

\subsubsection{Effect of Denoising Steps} We tune the numbers of denoising steps of four scales ($i.e.$, $T_{1},T_{2},T_{3}$ and $T_{4}$) to explore their optimal configurations. Specifically, as shown in Table \ref{tab6}, we first balance the number of denoising steps on four scales and provide three options for them: $100$, $250$, and $1000$. Then, we further deploy two unbalanced denoising strategies on four scales. Analyzing the results shown in Table \ref{tab6}, we can see that the configuration of $T_{1}=T_{2}=T_{3}=T_{4}=1000$ obtains the best generation performance, considering the additional computional time cost, its performance gains are limited. Finally, we set the number of denoising steps on four scales as $250$. Notably, Table \ref{tab6} further indicates that the cooperative condition input of text embedding and coarser motion embedding benefits better denoising quality with fewer denoising steps, significantly facilitating the denoising diffusion process. 

\begin{table}[t]
	\def\arraystretch{1.35}
	\resizebox{0.49\textwidth}{!}{
		\begin{tabular}{cccc|cccc}
			\toprule
			\multicolumn{4}{c|}{Number of Denoising Steps} & \multirow{2}{*}{\begin{tabular}[c]{@{}c@{}}R-Precision \\ Top-1 \end{tabular} $\uparrow$} & \multirow{2}{*}{FID $\downarrow$} & \multirow{2}{*}{Diversity $\uparrow$} & \multirow{2}{*}{Time $\downarrow$} \\
			\cline{1-4}
			\; $T_{1}$ \;   & \; $T_{2}$ \;   & \; $T_{3}$ \;   & \; $T_{4}$ \;    &                                                                      &                      &                            &                           \\ \hline           
			$100$ &  $100$   & $100$  & $100$      & $0.473$ & $0.362$ & $9.696$  & $0.8$ \\
			$250$ &  $250$   & $250$  & $250$     & $0.503$ & $0.109$ & $\boldsymbol{9.826}$  & $1.3$ \\
			$1000$ &  $1000$   & $1000$  & $1000$  & $\boldsymbol{0.505}$ & $\boldsymbol{0.108}$ & $9.805$  & $6.8$ \\
			$1000$ &  $1000$   & $500$  & $500$      & $0.504$ & $0.109$ & $9.829$  & $3.7$ \\
			$500$ &  $500$   & $1000$  & $1000$      & $0.504$ & $0.109$ & $9.828$  & $3.7$ \\
			
			\bottomrule
		\end{tabular}
	}
	\vspace{1.5mm}
	\caption{The performance comparison between different number of  denoising steps on each scale. The time performance we reported is the average inference time (second) of each sentence.}
	\label{tab6}
\end{table}

\section{Limitation and Future Work}
In this section, we analyze the limitation of GUESS to inspire its further development. We preliminarily summarize GUESS's further development into two aspects. Firstly, we consider our multi-stage scheme in the current version to be a static network that deploys fixed four pose scales and four inference stages on all input text samples. In other words, its number of inference stages is sample-independent and fixed across all input textual descriptions. In future work, we will develop it into a dynamic one that can adaptively adjust its number of inference stages based on the different text description inputs. 

Second, we can further develop the motion guess from the spatial dimension to the temporal dimension. Specifically, we can generate a human motion sequence of increasing temporal resolution by inferring a low-temporal-resolution guess first and then successively adding higher-temporal-resolution details. As an initial attempt at progressive text-to-motion generation, we hope GUESS can inspire more investigation and exploration in the community.

\section{Conclusion}
In this paper, we propose GUESS, a powerful progressive generation strategy for the text-driven human motion synthesis task. Firstly, we represent a human pose with a multi-scale skeleton and stabilize its motion at multiple abstraction levels. Then, we deploy a VAE on each pose scale to learn its latent motion embedding. Finally, a cascaded latent diffusion model facilitates the probabilistic text-to-motion mapping with cooperative guidance of textual embedding and gradually richer motion embedding. Besides, we further integrate GUESS with the proposed dynamic multi-condition fusion mechanism to dynamically balance the cooperative effects of the given textual condition and synthesized coarse motion prompt in each input sample and generation stage. Extensive experiments verify that GUESS outperforms previous state-of-the-art methods on large-scale datasets, developing a strong baseline for high-quality and diverse generation.

{\small
	\bibliographystyle{ieee_fullname}
	\bibliography{main}
}

\begin{IEEEbiography}[{\includegraphics[width=1in,height=1.25in,clip,keepaspectratio]{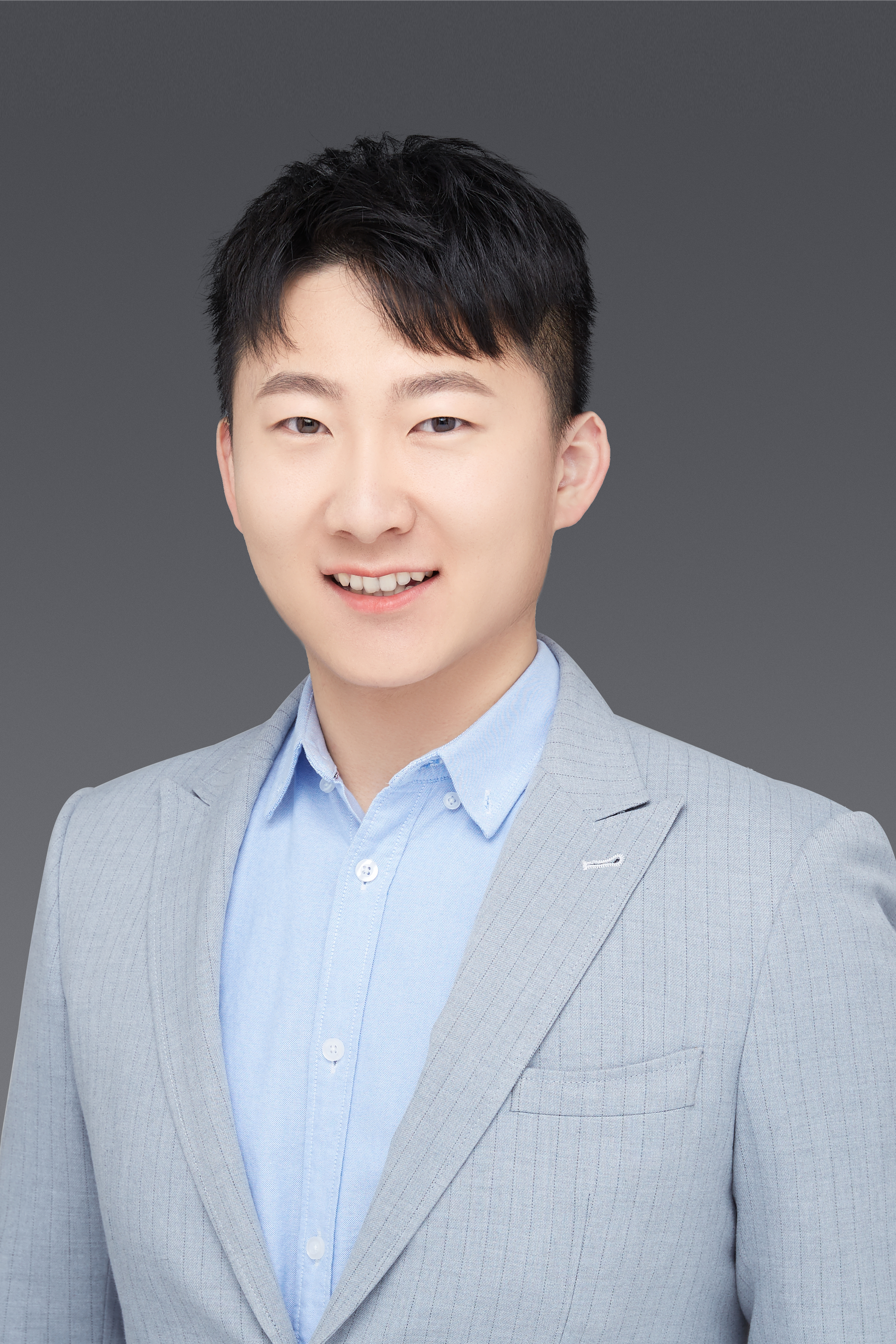}}]{Xuehao Gao}
	is a 2nd year Ph.D. student at Xi'an Jiaotong University. He is working toward the Ph.D. degree in control science and engineering at national key laboratory of human-machine hybrid augmented intelligence. His research interests include graph representation learning, human action recognition, prediction and synthesis. He has published papers in CVPR, IEEE-TNNLS, IEEE-TMM and IEEE-TCSVT ect. He is a student member of the IEEE.
\end{IEEEbiography}
\vspace{-0.5in}
\begin{IEEEbiography}[{\includegraphics[width=1in,height=1.25in,clip,keepaspectratio]{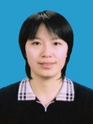}}]{Yang Yang}
	received B.E. degree in Information Engineering from Xi'an Jiaotong University, China, in 2005, and the Double-degree Ph.D in Pattern Recognition and Intelligent System from Xi'an Jiaotong University, China, and Systems Innovation Engineering from Tokushima University, Japan, in 2011. She is currently an Associate Professor of the School of Electronic and Information Engineering, Xi'an Jiaotong University, China. Her research interests include image processing, multimedia and machine learning.
\end{IEEEbiography}
\vspace{-0.5in}
\begin{IEEEbiography}[{\includegraphics[width=1in,height=1.25in,clip,keepaspectratio]{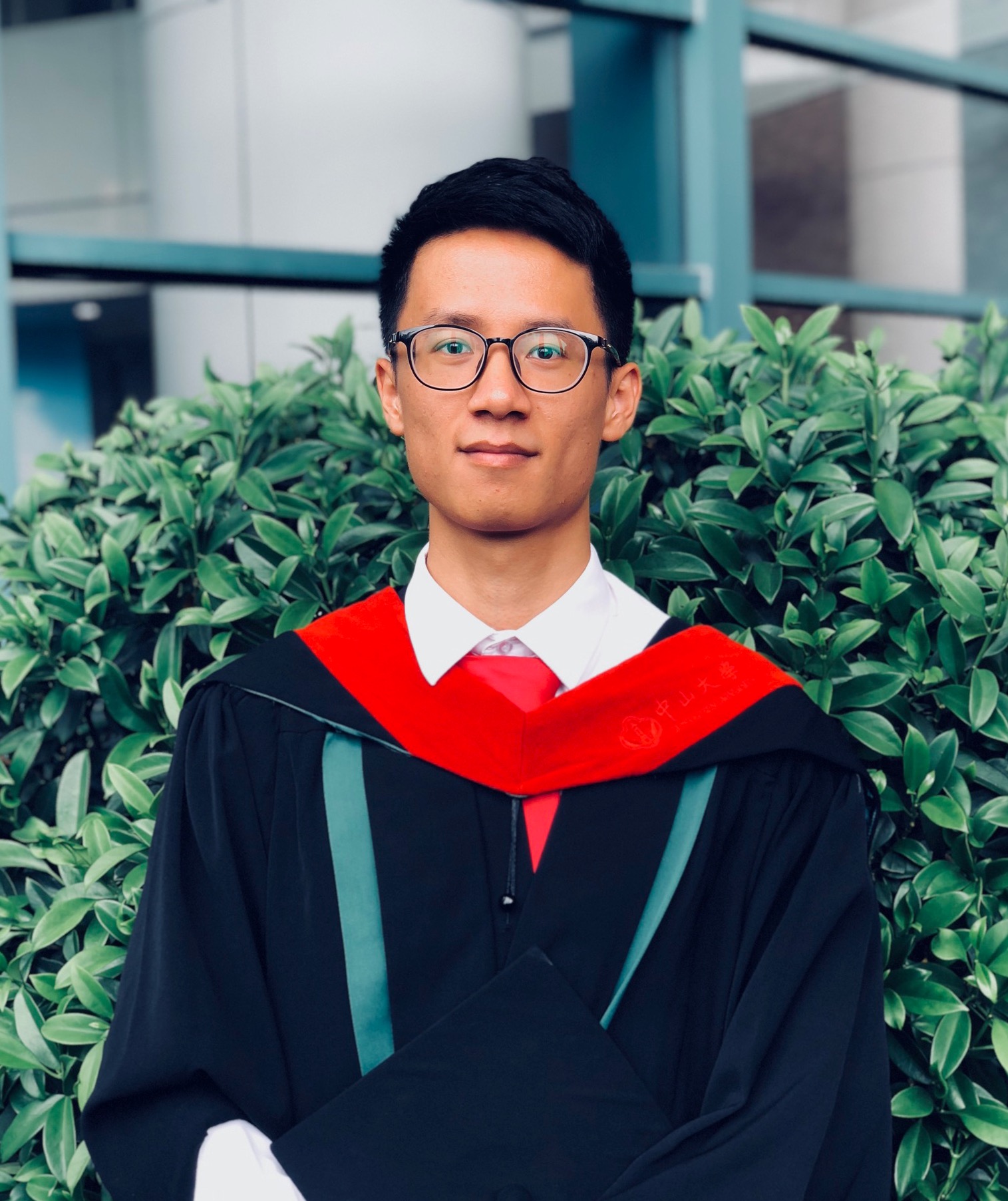}}]{Zhenyu Xie}
	received his bachelor degree from Sun Yat-sen University and is studying for the Ph.D. degree in the School of Intelligent Systems Engineering at Sun Yat-sen University. Currently, his research interests mainly lie in the Human-centric Synthesis, including but not limited to 2D/3D virtual try-on, 2D/3D-aware human synthesis/editing, cross-modal human motion synthesis, cross-modal video synthesis, etc.
\end{IEEEbiography}
\vspace{-0.5in}
\begin{IEEEbiography}[{\includegraphics[width=1in,height=1.25in,clip,keepaspectratio]{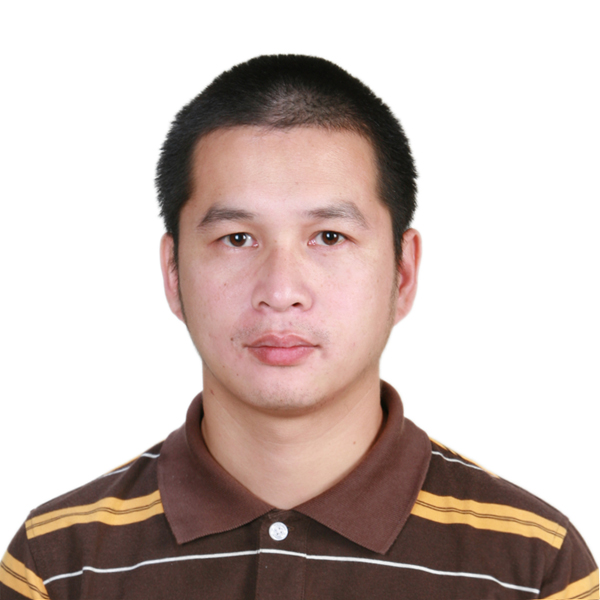}}]{Shaoyi Du}	
	received B.S. degrees both in computational mathematics and in computer science, M.S. degree in applied mathematics and Ph.D. degree in pattern recognition and intelligence system from Xi’an Jiaotong University, China in 2002, 2005 and 2009 respectively. He worked as a postdoctoral fellow in Xi’an Jiaotong University from 2009 to 2011 and visited University of North Carolina at Chapel Hill from 2013 to 2014. He is currently a professor of Institute of Artificial Intelligence and Robotics in Xi’an Jiaotong University.
\end{IEEEbiography}
\vspace{-0.5in}
\begin{IEEEbiography}[{\includegraphics[width=1in,height=1.25in,clip,keepaspectratio]{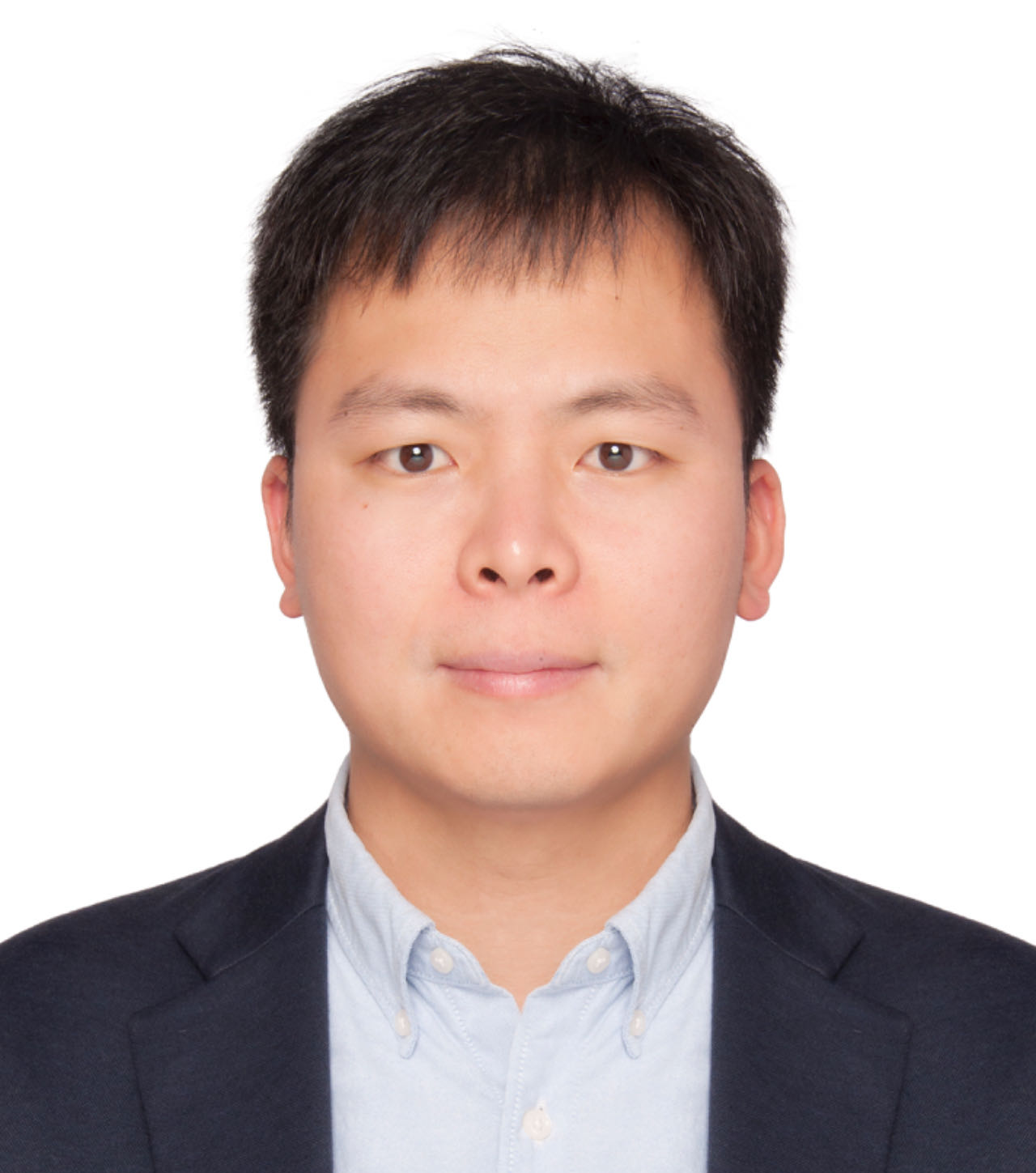}}]{Zhongqian Sun}	
	received the B.A. and M.S. degree in computer science and technology in 2009 and 2011 from Harbin Institute of Technology. He is now the Director of Tencent AI Lab. His primary research interests lie in 3D object reconstruction, character animation generation and image generation, etc.
\end{IEEEbiography}
	\vspace{-0.5in}
\begin{IEEEbiography}[{\includegraphics[width=1in,height=1.25in,clip,keepaspectratio]{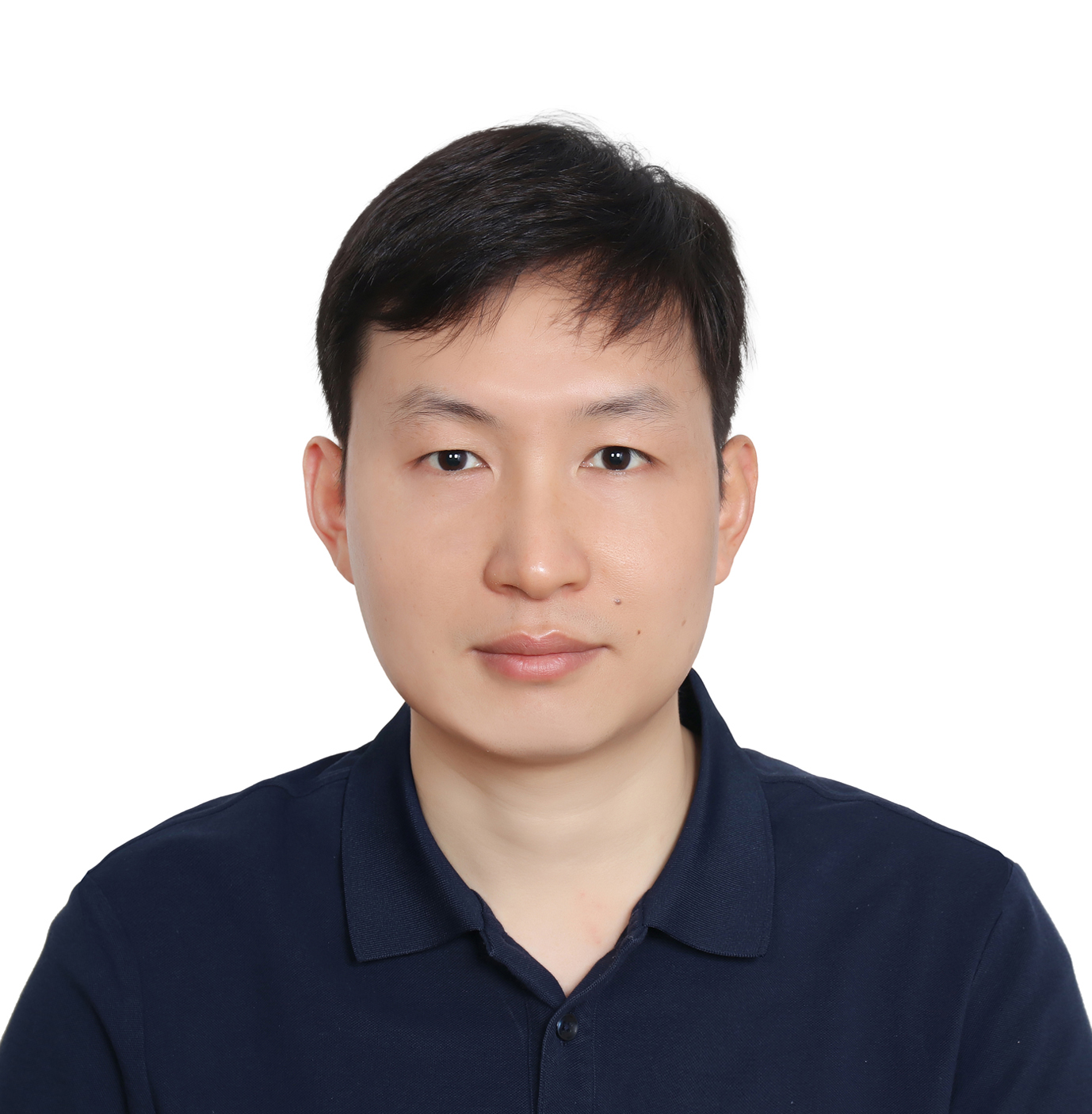}}]{Yang Wu}
	received a BS degree and a Ph.D. degree from Xi'an Jiaotong University in 2004 and 2010, respectively. He is currently a principal researcher with Tencent AI Lab. From Jul. 2019 to May 2021, he was a program-specific senior lecturer with the Department of Intelligence Science and Technology, Kyoto University. He was an assistant professor of the NAIST International Collaborative Laboratory for Robotics Vision, Nara Institute of Science and Technology (NAIST), from Dec.2014 to Jun. 2019. From 2011 to 2014, he was a program-specific researcher with the Academic Center for Computing and Media Studies, Kyoto University. His research is in the fields of computer vision, pattern recognition, as well as multimedia content analysis, enhancement and generation.
\end{IEEEbiography}

\end{document}